\documentclass[a4paper]{amsart}

\usepackage{amsmath,amssymb,amsthm}
\usepackage{fullpage}
\usepackage{mathtools}
\usepackage{tikz}
\usetikzlibrary{matrix,arrows}

\usepackage{graphicx}
\usepackage{subfigure}
\usepackage{wrapfig}
\usepackage{amsmath}
\usepackage{amssymb}
\usepackage{url}
\usepackage{color}
\usepackage{comment}
\usepackage{tikz}
\usetikzlibrary{arrows}
\usepackage{float}
\usetikzlibrary{shapes.geometric}
\usetikzlibrary{backgrounds}
\usepackage{enumitem}

\theoremstyle{plain}

\theoremstyle{definition}

\newcommand{\vr}[2]{\mathrm{VR}(#1;#2)}

\newcommand{\diam}{\mathrm{diam}}
\newcommand{\im}{\mathrm{im}}

\begin{document}

\title{A torus model for optical flow}
\author{Henry Adams}
\email{adams@math.colostate.edu}
\author{Johnathan Bush}
\email{bush@math.colostate.edu}
\author{Brittany Carr}
\email{carr@math.colostate.edu}
\author{Lara Kassab}
\email{kassab@math.colostate.edu}
\author{Joshua Mirth}
\email{mirth@math.colostate.edu}

\begin{abstract}
We propose a torus model for high-contrast patches of optical flow.
Our model is derived from a database of ground-truth optical flow from the computer-generated video \emph{Sintel}, collected by Butler et al.\ in \emph{A naturalistic open source movie for optical flow evaluation}.
Using persistent homology and zigzag persistence, popular tools from the field of computational topology, we show that the high-contrast $3\times 3$ patches from this video are well-modeled by a \emph{torus}, a nonlinear 2-dimensional manifold.
Furthermore, we show that the optical flow torus model is naturally equipped with the structure of a fiber bundle, related to the statistics of range image patches.
\end{abstract}


\maketitle

\section{Introduction}\label{sec:intro}

Video recordings collapse a moving three-dimensional world onto a two-dimensional screen.
How do the projected two-dimensional images vary with time?
This apparent two-dimensional motion is called {\em optical flow}.
More precisely, the optical flow at a frame in a video is a vector field, where the vector at each pixel points to where that pixel appears to move for the subsequent frame~\cite{beauchemin1995computation}.

Algorithms estimating optical flow must exploit or make assumptions about the prior statistics of optical flow.
Indeed, it is impossible to recover the optical flow vector field using only a video recording.
For example, if one is given a video of a spinning barber's pole, one does not know (a priori) whether the pole is moving up, or instead spinning horizontally.
The estimation of optical flow from a video sequence is a useful step in many computer vision tasks~\cite{barron1994performance,fleet2006optical}, including for example robotic motion.
Therefore there is a substantial interest in the statistics of optical flow.

\begin{figure}[htb]
\centering
\includegraphics[scale=0.1]{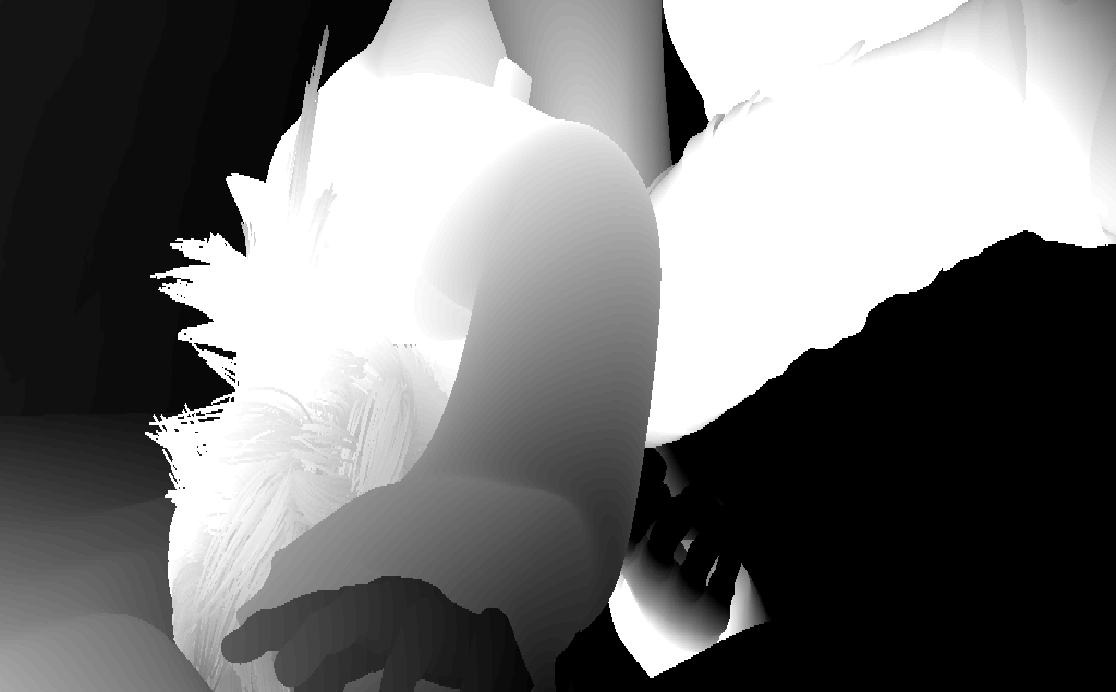}
\hspace{0.3mm}
\includegraphics[scale=0.1]{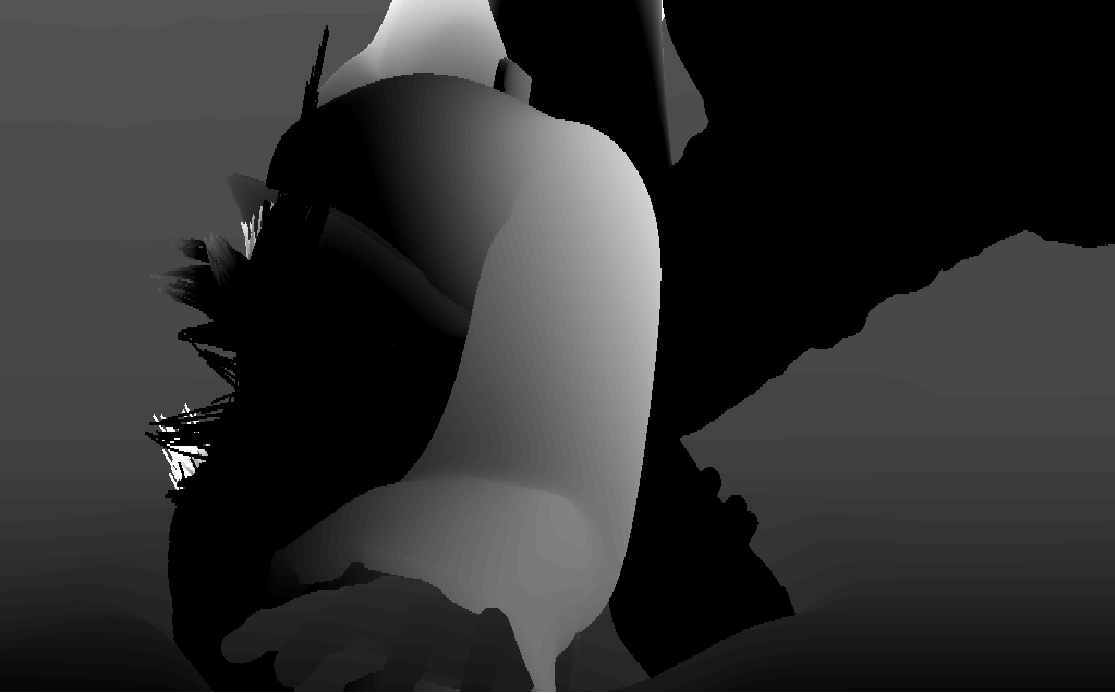}\\
\vspace{1mm}
\includegraphics[scale=0.1]{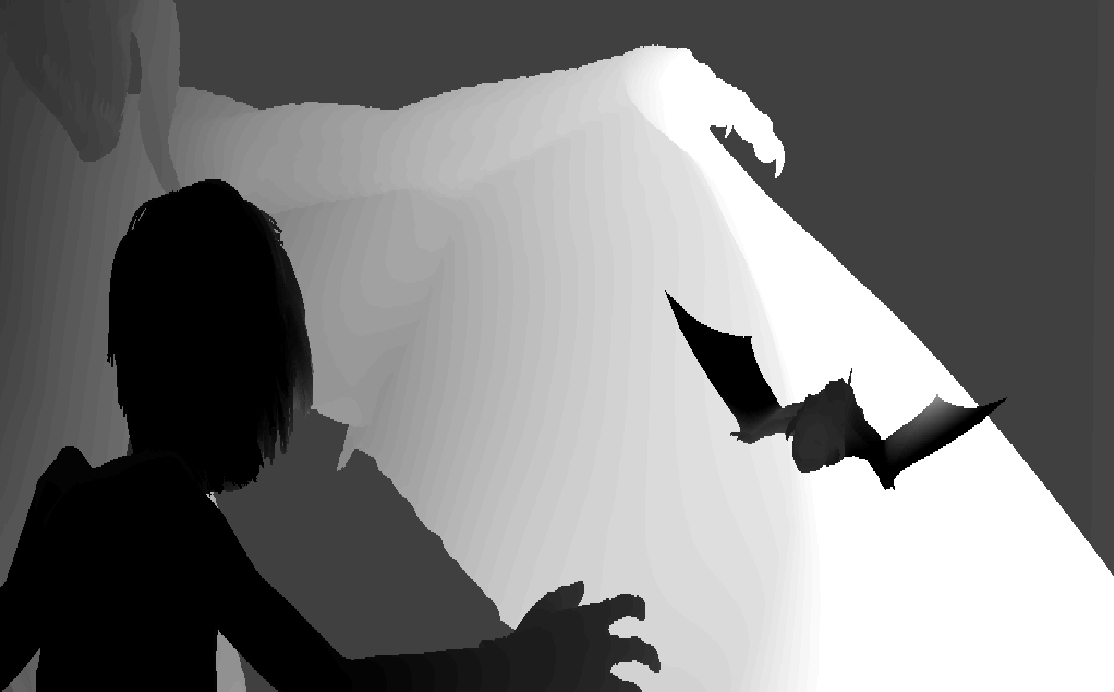}
\hspace{0.3mm}
\includegraphics[scale=0.1]{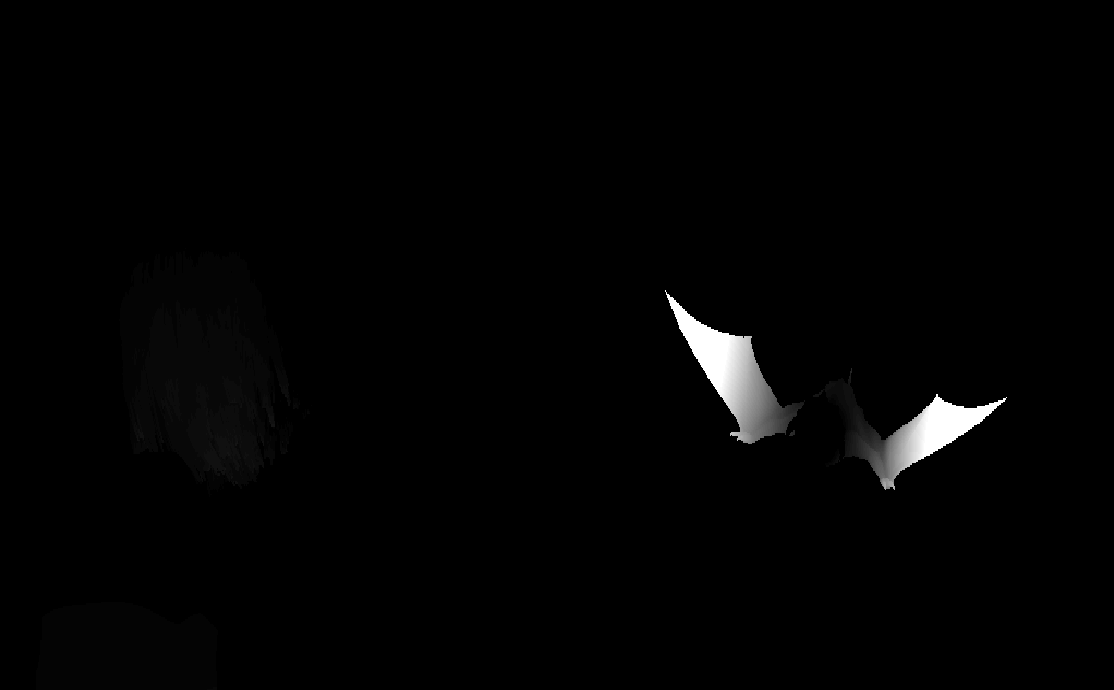}
\caption{Two sample optical flows extracted from the \emph{Sintel} database.
(Left column) horizontal components and (Right column) vertical components of optical flow.
White corresponds to flow in the positive direction ($+x$ or $+y$), and black corresponds to the negative direction.}
\label{fig:sampleFlow}
\end{figure}

One example database of optical flow is from the video short \emph{Sintel}, which is a computer generated 3D film.
The movements and textures in \emph{Sintel} are complex and the scenes are relatively long.
Furthermore, since the film is open source, the optical flow data is available for analysis (see Figure~\ref{fig:sampleFlow}), as described in detail by~\cite{butler2012naturalistic}.
As no instrument measures ground-truth optical flow, databases of optical flow must be carefully generated, and \emph{Sintel} is one of the richest such datasets.

We study the nonlinear statistics of a space of high-contrast $3\times 3$ optical flow patches from the \emph{Sintel} dataset using the topological machinery of~\cite{carlsson2008local} and~\cite{Range}.
We identify the topologies of dense subsets of this space using Vietoris--Rips complexes and persistent homology.
The densest patches lie near a circle, the \emph{horizontal flow circle}~\cite{NEBinTDA}.
In a more refined analysis, we select out the optical flow patches whose predominant direction of flow is in a small bin of angle values.
The patches in each such bin are well-modeled by a circle; each such circle is explained by the nonlinear statistics of range image patches.
Together these circles at different angles stitch together to form a torus model for optical flow.
We explain the torus model via the mathematical data of a \emph{fiber bundle}, and experimentally verify the torus using zigzag persistence.

One could use the torus model for the nonlinear statistics of optical flow for optical flow compression.
Indeed, a $3\times 3$ optical flow patch could be stored not as a list of 9 vectors in $\mathbb{R}^2$, but instead as:
\begin{itemize}[noitemsep]
\item an average flow vector in $\mathbb{R}^2$,
\item two real numbers parametrizing a patch on a 2-dimensional torus, and
\item  a $3\times 3$ collection of error vectors, whose entries will tend to be small in magnitude.
\end{itemize}
This is the first step towards a Huffman-type code~\cite{huffman1952method}, as used for example in JPEG~\cite{wallace1992jpeg}, and considered for optical images in~\cite{carlsson2008local} using a Klein bottle model.
\cite{perea2014klein} projects image patches to this Klein bottle and use Fourier-theoretic ideas to create a rotation-invariant descriptor for texture; perhaps similar ideas with projections to the flow torus could be used to identify different classes of optical flow (for example, flow from an indoor scene versus flow from an outdoor scene).  

In Section~\ref{sec:related} we overview prior work, in Section~\ref{sec:topo} we introduce our topological techniques, and in Section~\ref{sec:space} we describe the spaces of high-contrast optical flow patches.
We present our main results in Section~\ref{sec:res}.
Our code is available at \url{https://bitbucket.org/Cross_Product/optical_flow/}.
A preliminary version of this paper appeared in~\cite{adams2019nonlinear}.
We have added an orientation check to distinguish between the torus and the Klein bottle, and furthermore, we describe the sense in which the base space of our fiber bundle model arises from the (nonlinear) statistics of range image patches.

\section{Prior Work}\label{sec:related}

In the field of computer vision, a computer takes in visual data, analyzes the data via various statistics, and then outputs information or a decision based on the data.
Optical flow is commonly computed in computer vision tasks such facial recognition~\cite{bao2009liveness}, autonomous driving~\cite{kitti2013dataset}, and tracking problems~\cite{horn1981determining}.

Even though no instrument measures optical flow, there are variety of databases that have reconstructed ground truth optical flow via secondary means.
The Middlebury dataset in~\cite{baker2011database} ranges from real stereo imagery of rigid scenes to realistic synthetic imagery; the database contains public ground truth optical flow training data along with sequestered ground truth data for the purpose of testing algorithms.
The data from~\cite{ucl2012opticalFlow} consists of twenty different synthetic scenes with the camera and movement information provided.
The KITTI Benchmark Suite~\cite{kitti2013dataset} uses a car mounted with two cameras to film short clips of pedestrians and cars; attached scanning equipment allows one to reconstruct the underlying truth optical flow for data testing and error evaluation.
The database by~\cite{roth2007spatial} does not include accompanying video sequences.
Indeed, Roth and Black generate optical flow for a wide variety of natural scenes by pairing camera motions with range images (a range image contains a distance at each pixel); the resulting optical flow can be calculated from the geometry of the static scene and of the camera motion.
Only static scenes (with moving cameras) are included in this database: no objects in the field of view move independently.
By contrast, the ground-truth \emph{Sintel} optical flow database by~\cite{butler2012naturalistic}, which we study in this paper, is computed directly by projecting the the 3-dimensional geometry underlying the film.

Foundational papers that have analyzed the statistics of optical images from the perspective of computational topology include~\cite{lee2003nonlinear}, which proposes a circular model for $3\times 3$ optical image patches, and~\cite{carlsson2008local}, which uses persistent homology to extend this circular model to both a three-circle model and a Klein bottle model for different dense core subsets.

\section{Methods}\label{sec:topo}

Using only a finite sampling from some unknown underlying space, it is possible to estimate the underlying space's topology using persistent homology, as done by~\cite{carlsson2008local} for optical image patches.
For more information on homology see~\cite{armstrong2013basic,Hatcher}, for introductions to persistent homology see~\cite{Carlsson2009,EdelsbrunnerHarer,edelsbrunner2000topological,zomorodian2005computing}, and for applications of persistent homology to sensor networks, machine learning, biology, medical imaging, see~\cite{PersistentImages,baryshnikov2009target,bendich2016persistent,bubenik2015statistical,chung2009persistence,lum2013extracting,Coordinate-free,topaz2015topological,xia2014persistent}.

First we thicken our finite sampling $X$ into a larger space giving an approximate cellularization of the unknown underlying space.
We use a \emph{Vietoris--Rips simplical complex} of $X$ at scale $r\ge 0$, denoted $\vr{X}{r}$.
The vertex set is some metric space (or data set) $(X,d)$, and $\vr{X}{r}$ has a finite subset $\sigma\subseteq X$ as a face whenever $\diam(\sigma)\leq r$ (i.e., whenever $d(x,x')\leq r$ for all vertices $x,x'\in\sigma$).
By definition, $\vr{X}{r} \subseteq \vr{X}{r'}$ whenever $r\leq r'$, so this forms a nested sequence of spaces as the scale $r$  increases.

\begin{figure}[htp]
\centering
\includegraphics[scale=0.23]{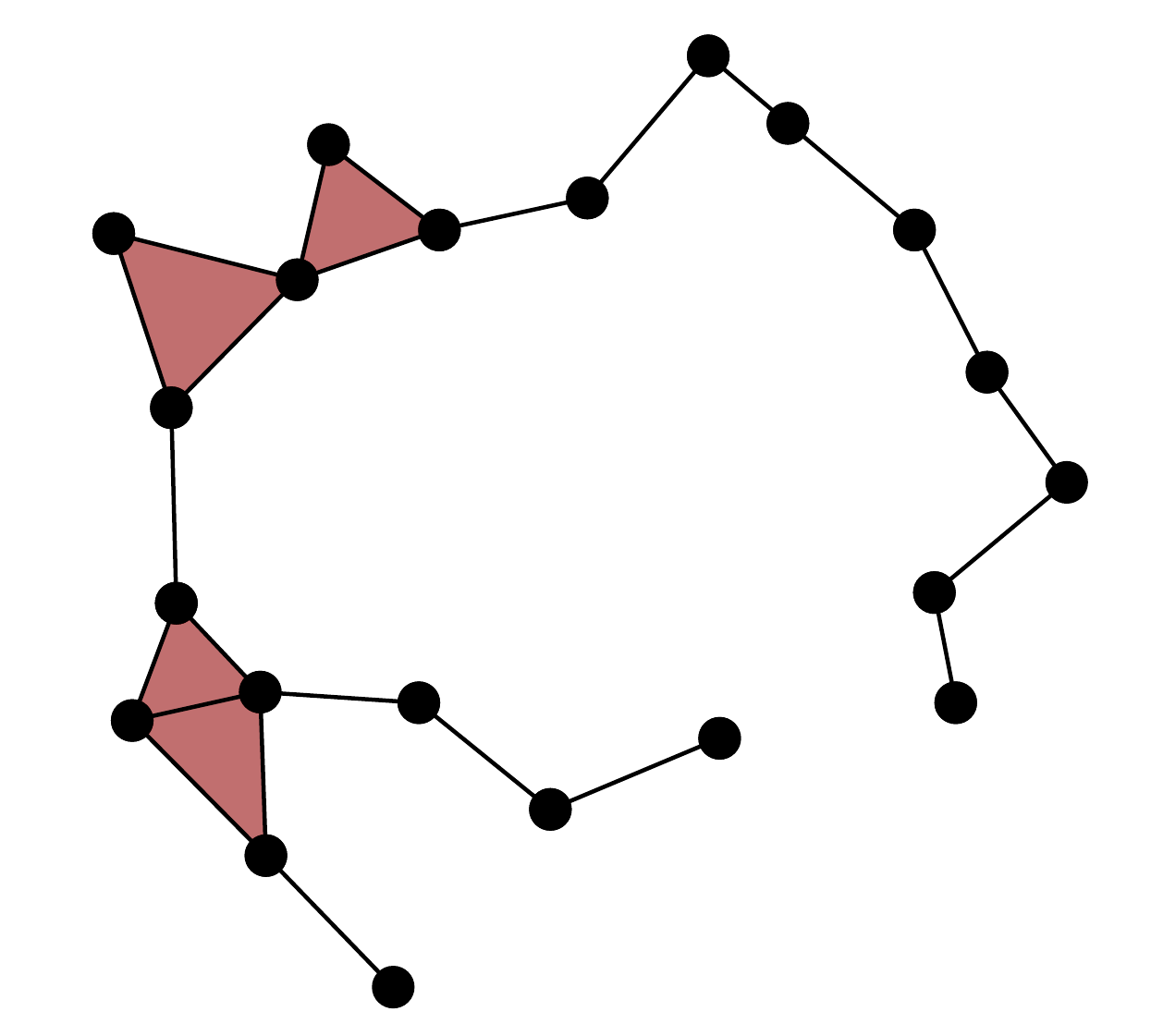}
\includegraphics[scale=0.23]{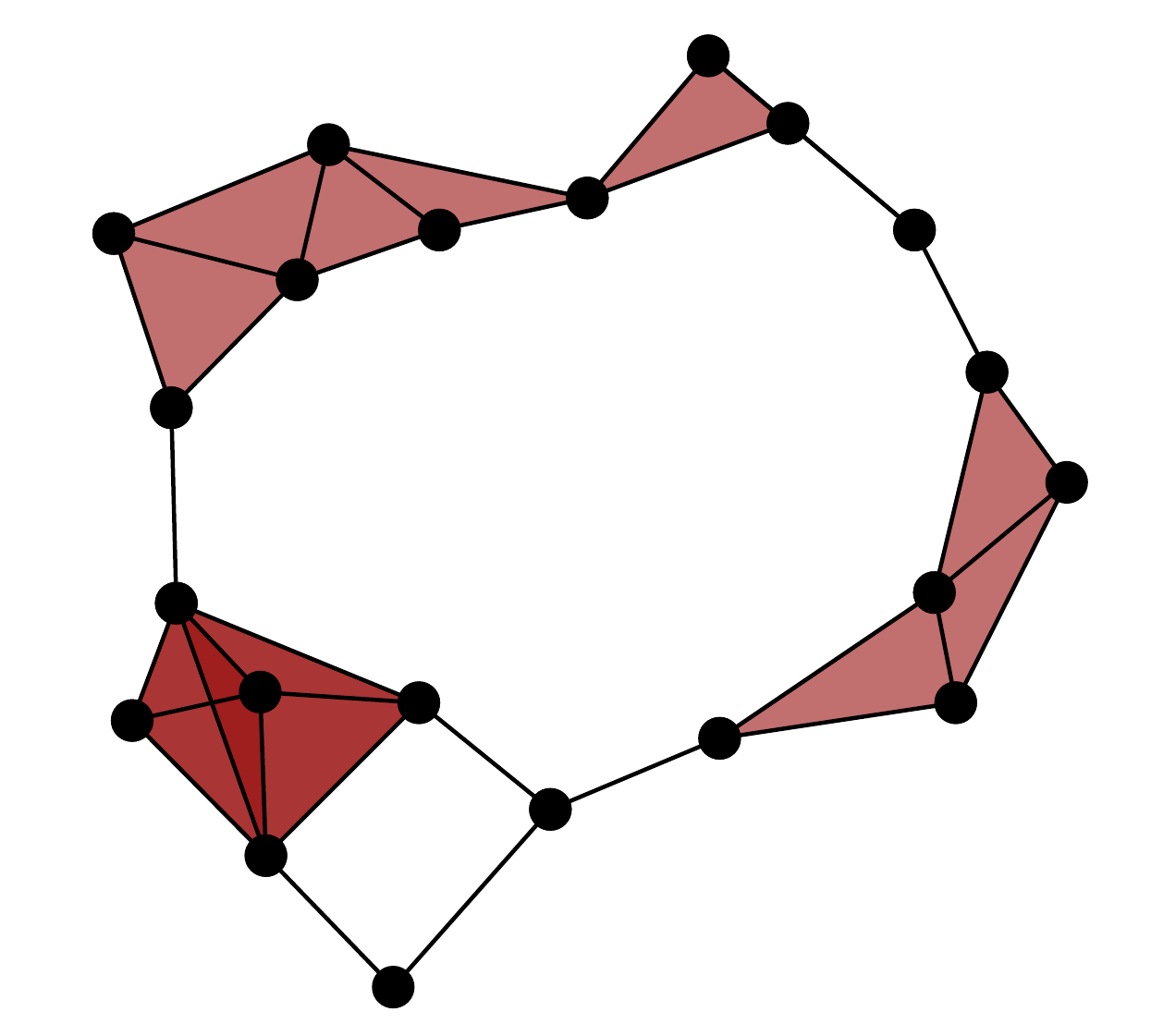}
\includegraphics[scale=0.23]{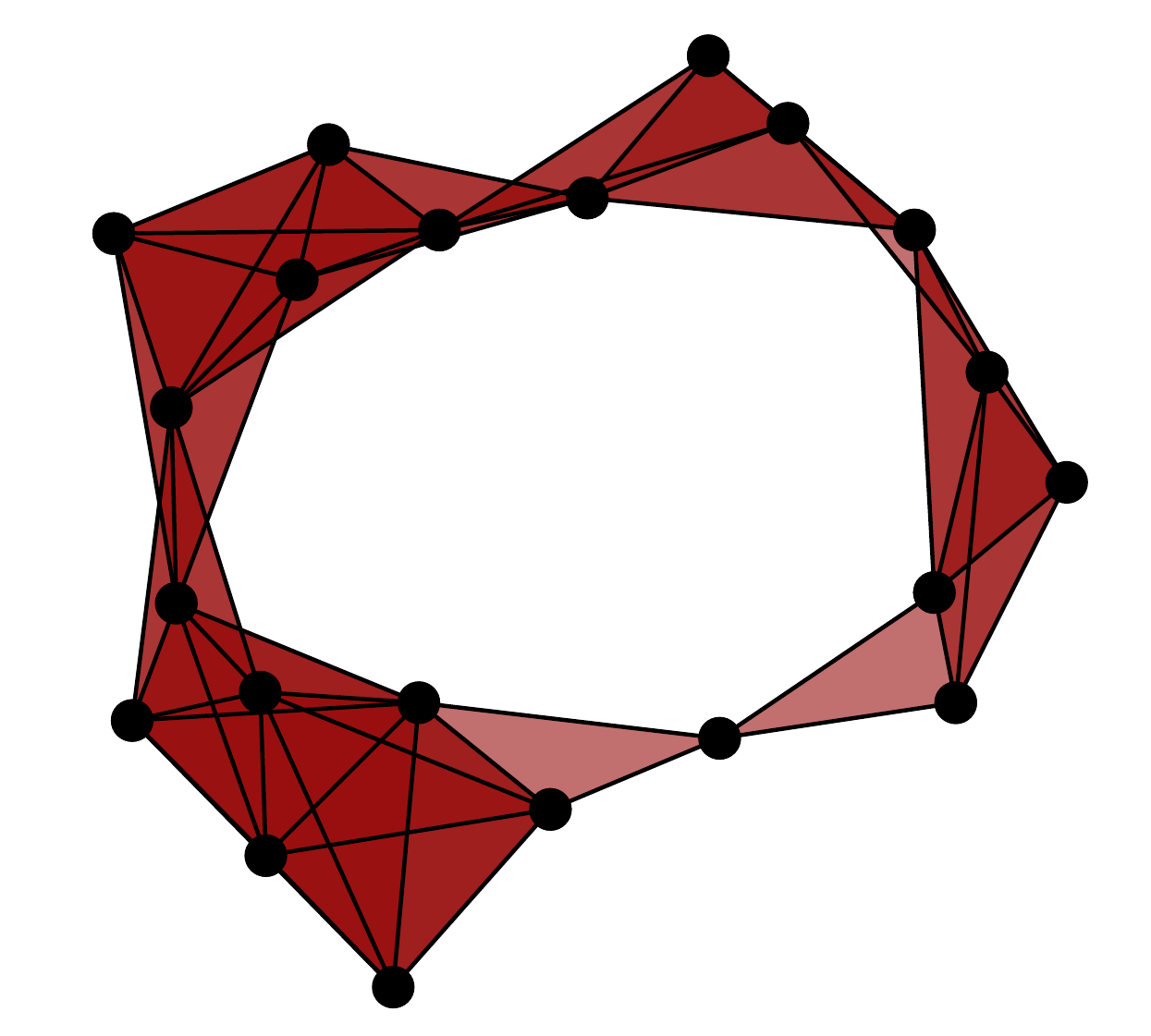}
\includegraphics[scale=0.23]{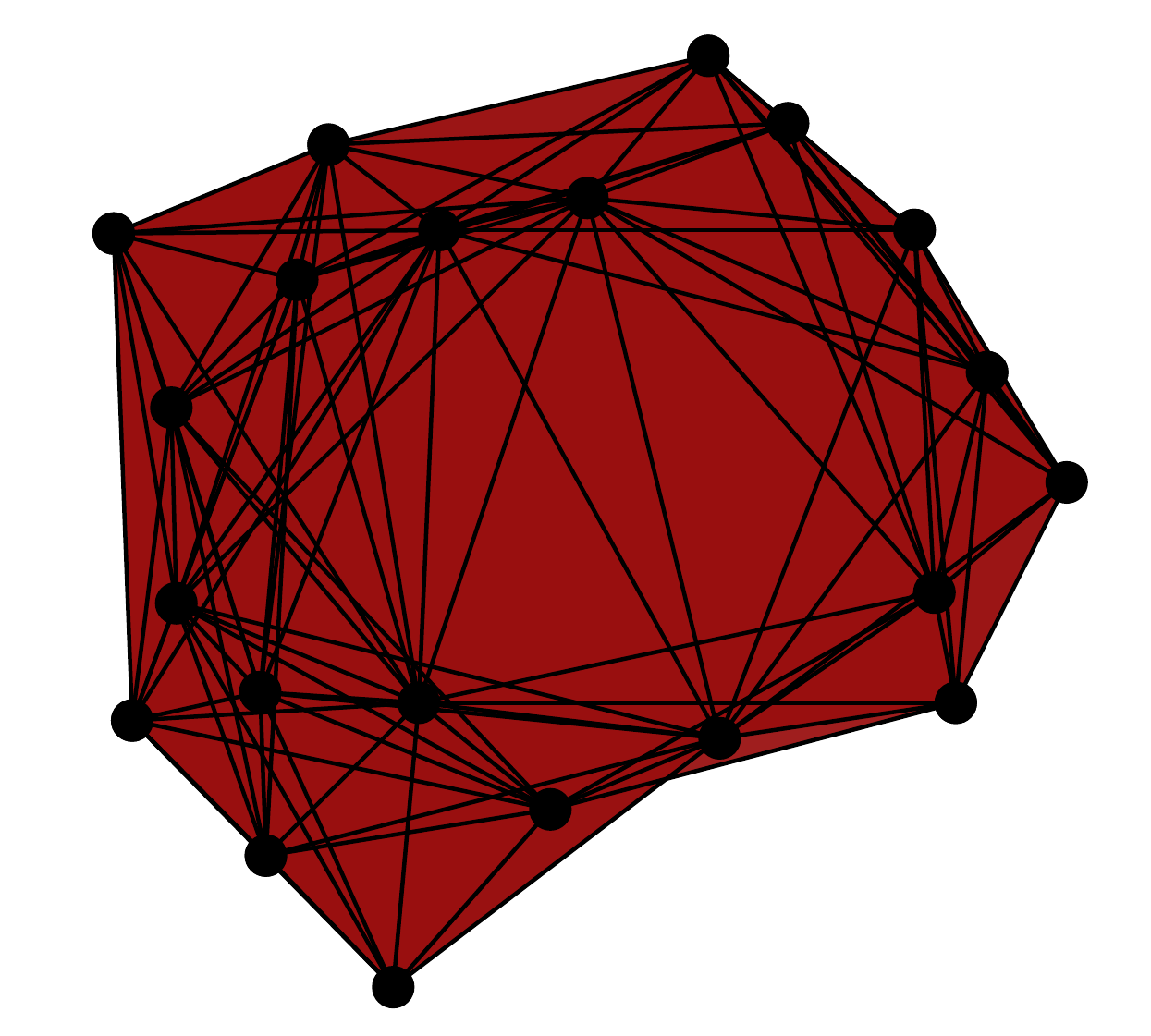}
\caption{Four nested Vietoris--Rips complexes, with $\beta_0$ equal to 1 in all four complexes, and with $\beta_1$ equal to 0, 2, 1, and 0.}
\label{fig:rips}
\end{figure}

For example, let $X$ be 21 points which (initially unknown to us) are sampled with noise from a circle.
Figure~\ref{fig:rips} contains four nested Vietoris--Rips complexes built from $X$, with $r$ increasing.
The black dots denote $X$.
At first $r$ is small enough that a loop has not yet formed.
As $r$ increases, we recover instead a figure-eight.
For larger $r$, $\vr{X}{r}$ recovers a circle.
Finally, $r$ is large enough that the loop has filled to a disk.

\begin{figure}[htp] 
\centering
\subfigure{\includegraphics[width=3.5in]{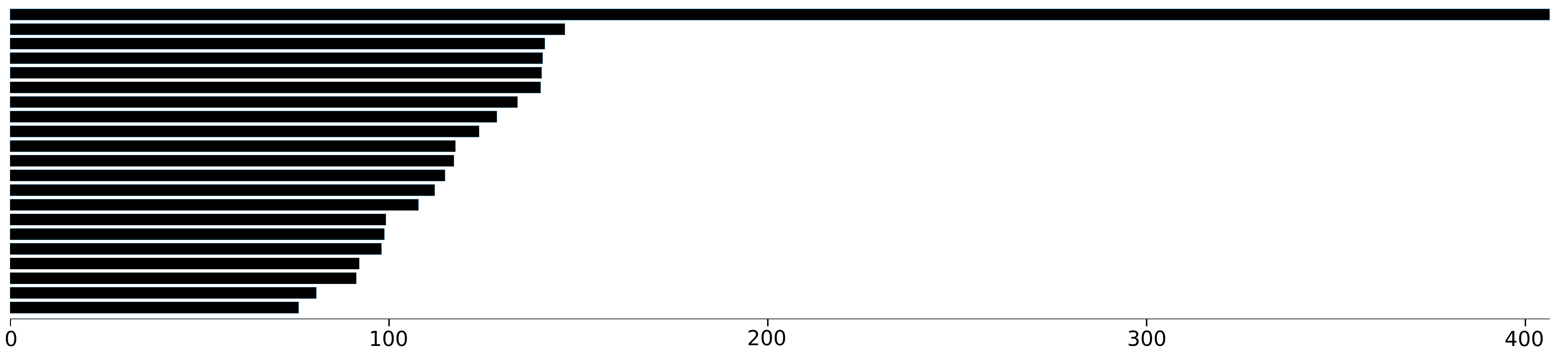}}
\subfigure{\includegraphics[width=3.5in]{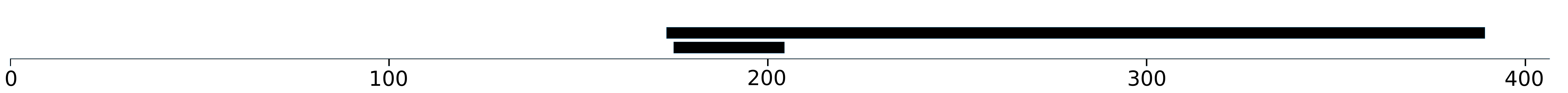}}
\caption{(Top) The $0$-dimensional persistence barcode associated to the dataset in Figure~\ref{fig:rips}.
(Bottom) The $1$-dimensional persistence barcode associated to the same dataset.}
\label{fig:circleBarcodes}
\end{figure}

Next we apply homology, an algebraic invariant.
The $k$-th Betti number of a topological space, denoted $\beta_k$, roughly speaking counts the number of ``$k$-dimensional holes" in a space.
More precisely, $\beta_k$ is the rank of the $k$-th homology group.
As a first example, the number of 0-dimensional holes in any space is the number of connected components.
For an $n$-dimensional sphere with $n\ge1$, we have $\beta_0=1$ (one connected component) and $\beta_n=1$ (one $n$-dimensional hole).

The choice of scale $r$ is important when attempting to estimate the topology of an underlying space by a Vietoris--Rips complex $\vr{X}{r}$ of a finite sampling $X$.
Indeed, without knowing the underlying space, we do not know how to choose the scale $r$.
We therefore use persistent homology~\cite{edelsbrunner2000topological,EdelsbrunnerHarer,zomorodian2005computing}, which allows us to compute the Betti numbers over a range of scale parameters $r$.
Persistent homology relies on the the fact that the map from a topological space $Y$ to its $k$-th homology group $H_k(Y)$ is a \emph{functor}: for $r\leq r'$, the inclusion $\vr{X}{r}\hookrightarrow \vr{X}{r'})$ of spaces induces a map $H_k\bigl(\vr{X}{r}\bigr)\to H_k\bigl(\vr{X}{r'}\bigr)$ between homology groups.

Figure~\ref{fig:circleBarcodes} displays persistent homology barcodes, with the horizontal axis encoding the varying $r$-values.
At a given scale $r$, the Betti number $\beta_k$ is the number of intervals in the dimension $k$ plot that intersect the vertical line through scale $r$.
The dimension $0$ plot shows $21$ disjoint vertices joining into one connected component as $r$ increases.
In the dimension $1$, the two intervals plot correspond to the two loops that appear: each interval begins when a loop forms and ends when that loop fills to a disk.
For a long range of $r $-values, the topological profile $\beta_0=1$ and $\beta_1=1$, is obtained.
Hence, this barcode reflects the fact that our points $X$ were noisily sampled from a circle.
Indeed, the idea of persistent homology is that long intervals in the persistence barcodes typically correspond to real topological features of the underlying space.

In zigzag persistence~\cite{carlsson2009zigzag,ZigzagPersistence}, a generalization of persistent homology, the direction of maps along a sequence of topological spaces is now arbitrary.
For example, given a large dataset $Y$, one may attempt to estimate the topology of $Y$ by instead estimating the topology of a number of smaller subsets $Y_i\subseteq Y$.
Indeed, consider the following diagram of inclusion maps between subsets of the data.
\begin{equation*}
Y_1\hookrightarrow Y_1 \cup Y_2 \hookleftarrow Y_2\hookrightarrow Y_2 \cup Y_3\hookleftarrow Y_3\hookrightarrow\cdots \hookleftarrow Y_n.
\end{equation*}
Applying the Vietoris--Rips construction at scale parameter $r$ and $k$-dimensional homology, we obtain an induced sequence of linear maps 
\begin{center}
\begin{tikzpicture}
\node at (-0.5, 0.45) (a) {$H_k\bigl(\vr{Y_1}{r}\bigr)$};
\node at (1, -0.45) (b) {$H_k\bigl(\vr{Y_1 \cup\, Y_2}{r}\bigr)$};
\node at (2.5, 0.45) (c) {$H_k\bigl(\vr{Y_{2}}{r}\bigr)$};
\node at (4, -0.45) (d) {$\cdots$};
\node at (5.5, 0.45) (e){$H_k\bigl(\vr{Y_n}{r}\bigr)$};
\draw [->] (a) -- (b);
\draw [->] (c) -- (b);
\draw [->] (c) -- (d);
\draw [->] (e) -- (d);
\end{tikzpicture}
\end{center}
which is an example of a \textit{zigzag diagram}.
Such a sequence of linear maps provides the ability to track features contributing to homology among the samples $Y_i$.
Generators for $H_k(\vr{Y_i}{r})$ and $H_k(\vr{Y_{i+1}}{r})$ which map to the same generator of $H_k\bigl(\vr{Y_i \cup\, Y_{i+1}}{r}\bigr)$ indicate a feature common to both $Y_i$ and $Y_{i+1}$.
By tracking features common to all samples $Y_i$, we can estimate the topology of $Y$ without explicitly computing the persistent homology of the entire dataset.

\section{Experiments on Spaces of Flow Patches}\label{sec:space}

\textit{Sintel}~\cite{roosendaal2010} is an open-source computer animated film containing a variety of realism-enhancing effects, including widely-varied motion, illumination, and blur.
The MPI-\emph{Sintel} optical flow dataset~\cite{butler2012naturalistic} contains $1041$ optical flow fields from 23 indoor or outdoor scenes in this film.
Each flow field is $1024\times 436$ pixels, and scenes are up to 49 frames long.

In similar preprocessing steps to those done by~\cite{Range,carlsson2008local,lee2003nonlinear}, we create two types of spaces of high-contrast optical flow patches, $X(k,p)$ and $X_\theta(k,p)$.
The version $X_\theta(k,p)$ includes only those optical flow patches whose predominant angle is near $\theta\in[0,\pi)$.

Step 1: From the MPI-\emph{Sintel} database, we choose a random set of $4\cdot10^5$ optical flow patches of size $3\times3$.
Each patch is a matrix of ordered pairs, where we denote by $u_i$ and $v_i$ the horizontal and vertical components of the flow vector at pixel $i$, arranged as follows.
\[\begin{bmatrix}
(u_1,v_1) & (u_4,v_4) & (u_7,v_7)\\
(u_2,v_2) & (u_5,v_5) & (u_8,v_8)\\
(u_3,v_3) & (u_6,v_6) & (u_9,v_9)
\end{bmatrix}\]
We rearrange each patch $x$ to be a length-18 vector, \\ $x=(u_1, \ldots, u_9, v_1, \ldots, v_9)^T \in \mathbb{R}^{18}$.
Let $u$ and $v$ to be the vectors of horizontal and vertical flow, namely $u=(u_1, u_2,\ldots, u_9)^T$ and $v=(v_1, v_2,\ldots, v_9)^T$.

Step 2: We compute the contrast norm $\|x\|_D$ for each patch $x$ by summing the squared differences between all adjacent pixels and then taking the square root:
\begin{align*}
\|x\|_D^2 &=\sum_{i \sim j} \|(u_i,v_i)-(u_j,v_j)\|^2
\\ & =\sum_{i \sim j}(u_i-u_j)^2+(v_i-v_j)^2 =u^TDu+v^TDv.
\end{align*}
Here $i\sim j$ denotes that pixels $i$ and $j$ are adjacent in the $3\times 3$ patch, and $D$ is a symmetric positive-definite $9\times9$ matrix that stores the adjacency information of the pixels in a $3\times 3$ patch~\cite{lee2003nonlinear}.

Step 3: We study only high-contrast flow patches, which we expect to follow a different distribution than low-contrast patches.
Indeed, we select those patches that have a contrast norm among the top 20\% of the entire sample.
We replace each selected patch $x$ with its contrast-normalized patch $x/\|x\|_D$, mapping each patch onto the surface of an ellipsoid.
Dividing by contrast norm zero is not a concern, as such patches are not high-contrast.

Step 4: We further normalize the patches to have zero average flow.
We replace each contrast-normalized vector $x$ with $(u_1-\bar{u}, \ldots, u_9-\bar{u}, v_1-\bar{v}, \ldots, v_9-\bar{v})^T$, where $\bar{u}=\frac{1}{9}\sum_{i=1}^{9}u_i$ is the average horizontal flow, and $\bar{v}=\frac{1}{9}\sum_{i=1}^{9}v_i$ is the average vertical flow.
The significance of studying mean-centered optical flow patches is that one can represent any optical flow patch as its mean vector plus a mean-centered patch.

Step 5: In the case of $X_\theta(k,p)$ (as opposed to $X(k,p)$), we compute the predominant direction of each mean-centered flow patch, as follows.
For each $3 \times 3$ patch, construct a $9 \times 2$ matrix $X$ whose $i$-th row is $(u_i , v_i)\in\mathbb{R}^2$.
Apply principal component analysis (PCA) to $X$ to retrieve the principal component with the greatest component variance (i.e., the direction that best approximates the deviation from the mean).
We define the \emph{predominant direction} of this patch to be the angle of this direction (in $[0,\pi)$ or $\mathbb{RP}^1$) .
Select only those patches whose predominant direction is in the range of angles from $\theta-\frac{\pi}{12}$ to $\theta+\frac{\pi}{12}$.

Step 6: If we have more than 50,000 patches, then for the sake of computational feasibility we subsample down to 50,000 random patches.

Step 7: At this stage we have at most 50,000 high-contrast normalized optical flow patches.
We restrict to dense core subsets thereof, instead of trying to approximate the topology of such a diverse space.
We use the density estimator $\rho_k$, where $\rho_k(x)$ is the distance from $x$ to its $k$-th nearest neighbor; $\rho_k$ is inversely proportional to density.
We obtain a more local (or global) estimate of density by decreasing (or increasing) the choice of $k$.
Based on the density estimator $\rho_k$, we select out the top $p\%$ densest points.
We denote this set of patches by $X(k,p)$ (or $X_\theta(k,p)$ in the case where Step~5 is performed).

Some remarks on the preprocessing steps are in order.
Studying only high-contrast patches (Step~3) prevents the dataset from being a ``cone" (with apex a constant gray patch), and hence contractible.
The dataset is still extremely high-dimensional, however.
Dividing by the contrast norm in Step~3 maps the data from $\mathbb{R}^{18}$ to a 17-dimensional sphere thereof, and Step~4 maps the data to a 15-dimensional sphere.
Though the normalizations are important for our analysis, the normalized data is still 15-dimensional, and hence it is not at all clear that we will succeed in Section~\ref{sec:res} in finding 1- and 2-dimensional models for dense core subsets of this 15-dimensional data.
Our models will only be for dense core subsets of the data, which are produced via density thresholding in Step~5.
Some optical flow patches, such as those created by zooming in or zooming out on a flat wall, are certainly present in the \emph{Sintel} dataset, but not with high-enough frequency to remain after the density thresholding in Step~5 nor to appear in our 1- and 2-dimensional models in Section~\ref{sec:res}.

Though \emph{Sintel} is one of the richest optical flow datasets, we emphasize that it is created synthetically, and to some degree its statistics will vary from the optical flow in real-life videos.
We would be interested in the patterns that arise in larger (say $5\times 5$ or $7\times 7$) patches, though in this paper we restrict attention to $3\times 3$ patches.

\section{Results and Theory}\label{sec:res}

\begin{figure}[htp]
\centering
\includegraphics[height=2.5cm]{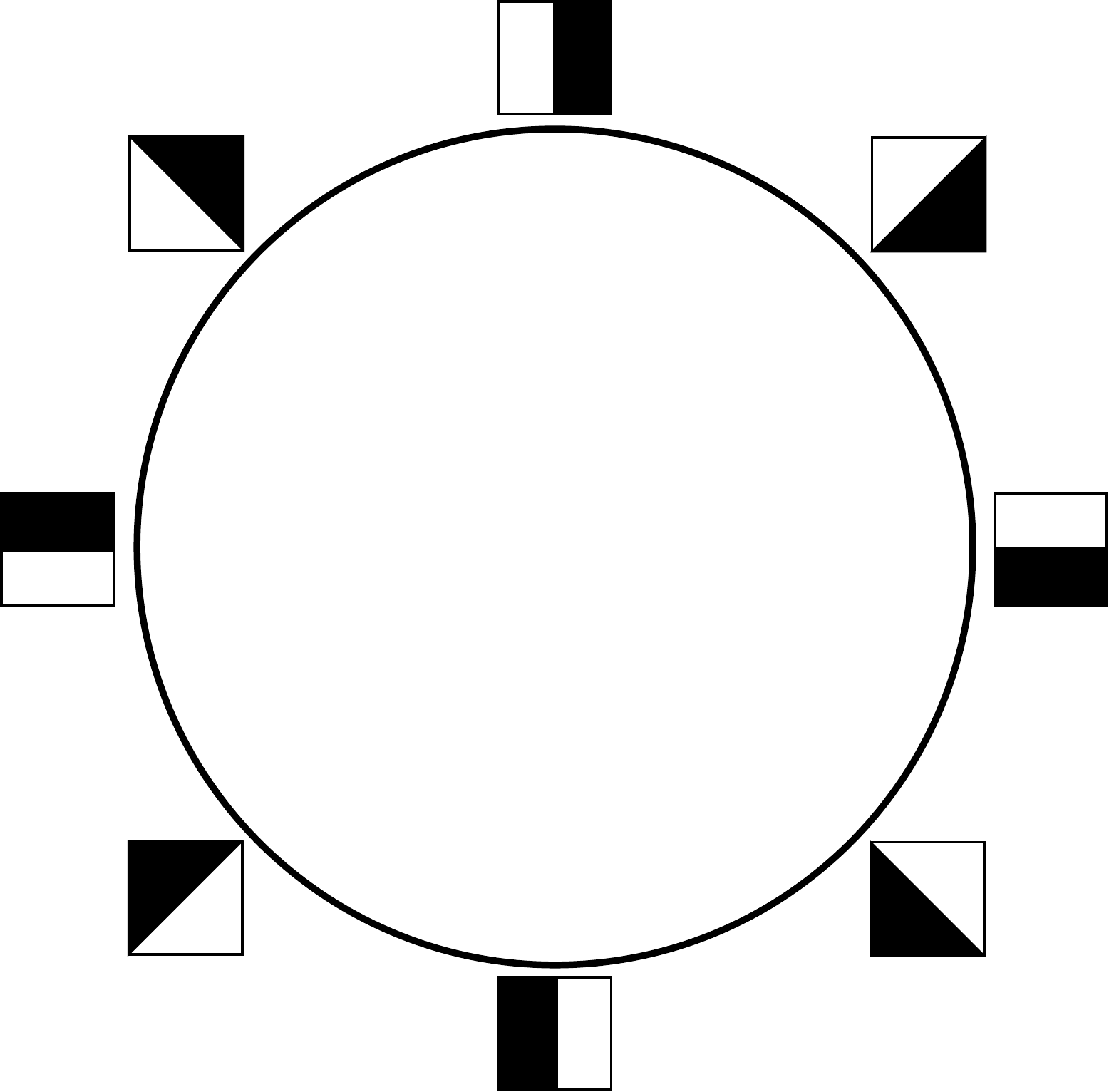}
\hspace{1mm}
\includegraphics[scale=0.15]{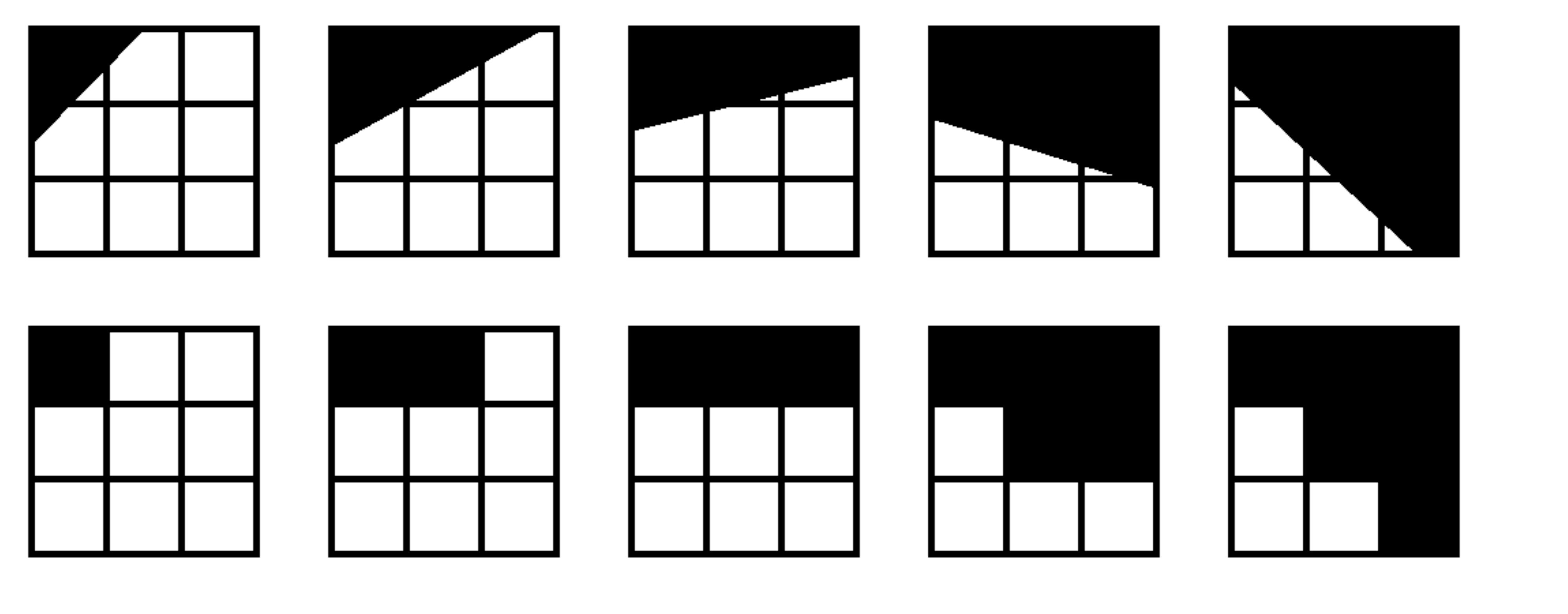}
\caption{(Left) Range patch primary circle.
White regions are background; black regions are foreground.
(Right) The top row contains linear step edges; the bottom row contains their range image binary approximations.}
\label{fig:primBin}
\end{figure}

Before describing our results on optical flow patches, we first describe the nonlinear statistics of range image patches (which contain a distance at each pixel), which will play an important role in the theory behind our results.
\cite{lee2003nonlinear} observes that high-contrast $3\times3$ range patches from~\cite{huang2000statistics} cluster near binary patches.
\cite{Range} uses persistent homology to find that the densest range clusters are arranged in the shape of a circle.
After enlarging to $5\times5$ or $7\times7$ patches, the entire primary circle in Figure~\ref{fig:primBin}(a) is dense.
The patches forming the range primary circle are binary approximations to linear step edges; see Figure~\ref{fig:primBin}(b).

\subsection{The horizontal flow circle}\label{ssec:hor}

\begin{figure}[htp]
\centering
\includegraphics[height=1.63in]{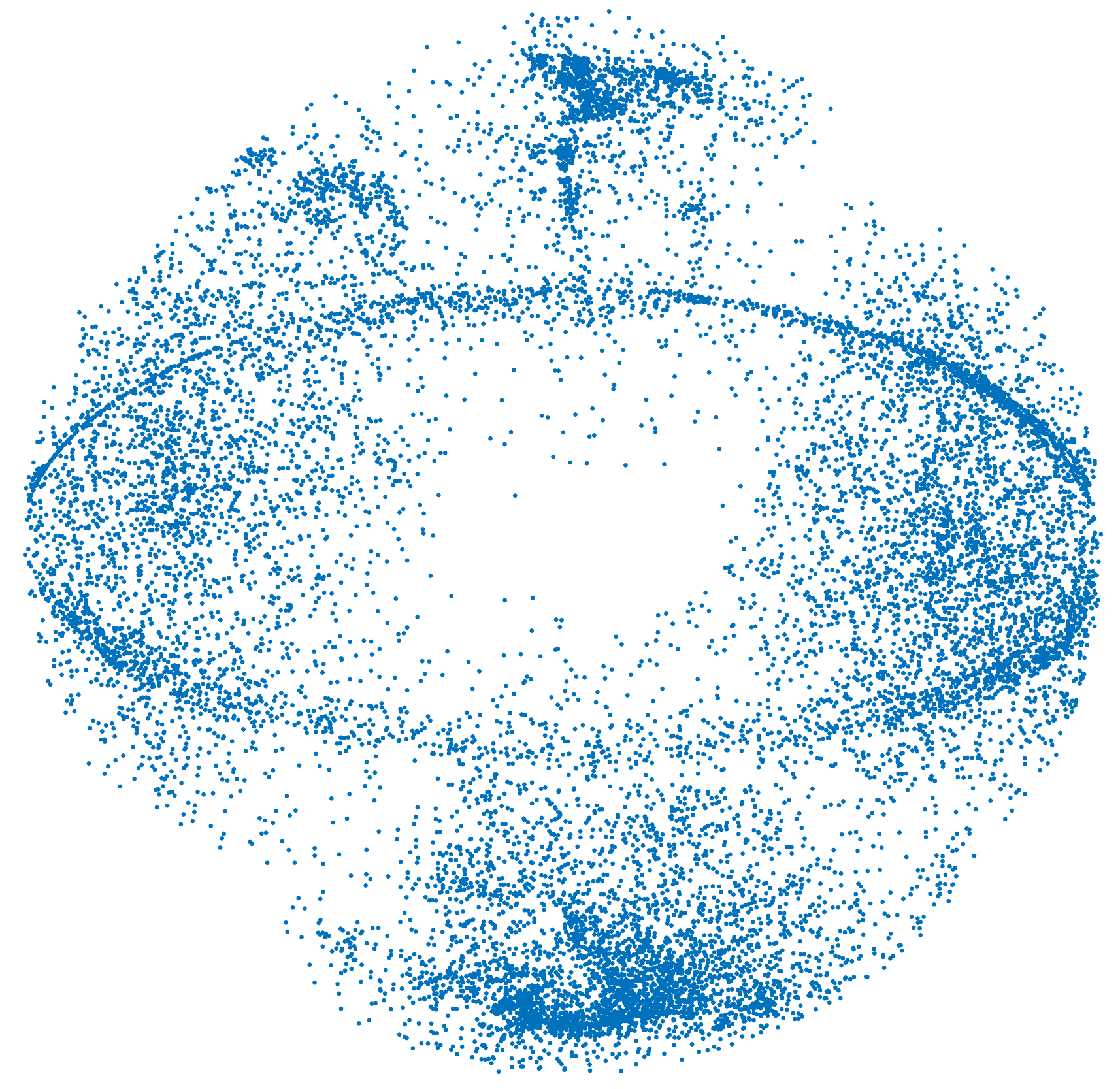}
\hspace{1mm}
\includegraphics[height=1.63in]{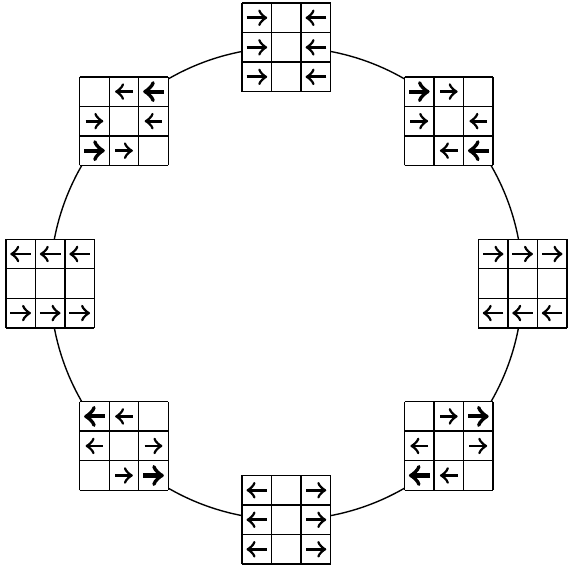}
\caption{
(Left) Projection of $X(300,30)$ onto $e_1^u$ and $e_2^u$.
(Right) The horizontal flow circle.
The patch at angle $\alpha$ is $\cos(\alpha)e_1^u+\sin(\alpha)e_2^u$.
}
\label{fig:gk300c30}
\end{figure}

\begin{figure}[htb]
\centering
\includegraphics[width=3in]{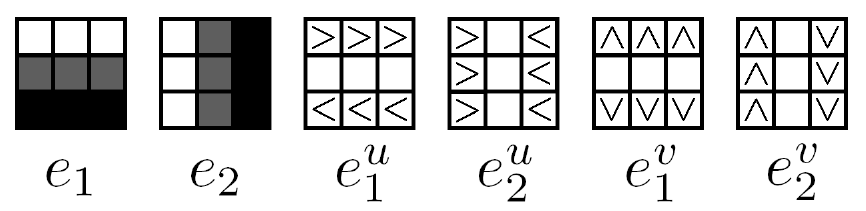}
\caption{In the $e_1$ and $e_2$ DCT patches, white pixels are positive and black negative.
The arrows in the flow patches $e_1^u$, $e_2^u$, $e_1^v$, and $e_2^v$ show the optical flow vector field patch.
}
\label{fig:dct}
\end{figure}

Using the the nudged elastic band method,~\cite{NEBinTDA} found that the dense core subset $X(300,30)$ is well-modeled by a horizontal flow circle.
We instead project onto suitable basis vectors in order to explain this circular model.
Let $e_1, e_2, \ldots, e_8$ be the discrete cosine transform (DCT) basis for $3\times3$ scalar patches, normalized to have mean zero and contrast norm one~\cite{lee2003nonlinear}.
We rearrange each $e_i$ to be a vector of length 9.
For each $i=1,2, \ldots,8$, we define optical flow vectors $e_i^u=\begin{pmatrix}e_i \\ \vec{0}\end{pmatrix}$ and $e_i^v=\begin{pmatrix}\vec{0} \\ e_i\end{pmatrix}$, where $\vec{0}\in\mathbb{R}^9$ is the vector of all zeros.
The vectors $e_i^u,e_i^v\in\mathbb{R}^{18}$ correspond respectively to optical flow in the horizontal and vertical directions; four of these basis vectors are in Figure~\ref{fig:dct}.
We change coordinates from the canonical basis for $\mathbb{R}^{18}$ to the 16 basis vectors $e_1^u, \ldots, e_8^u, e_1^v, \ldots,e_8^v$ (only 16 basis vectors are needed to model patches with zero average flow).
Projecting $X(300,30)$ onto basis vectors $e_1^u$ and $e_2^u$, as shown in Figure~\ref{fig:gk300c30}~(left), reveals the circular topology.

We denote by $S^1$ the interval $[0,2\pi]$ with its endpoints identified.
The patches in $X(300,50)$ lie near $\{\cos(\alpha)e_1^u+\sin(\alpha)e_2^u \ |\ \alpha\in S^1\}$, which we call the {\em horizontal flow circle}.
To explain why the horizontal circle is high-density, we will use the statistics of both the camera motion database and the range image database.

\begin{wrapfigure}{r}{0.6in}
\centering
\vspace{-\intextsep}
\includegraphics[width=0.6in]{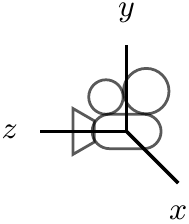}
\vspace{-2\intextsep}
\end{wrapfigure}
Any camera motion can be decomposed into six sub-motions: translation in the $x$, $y$, or $z$ direction (commonly referred to as right-left, up-down, or inward-outward translation), and rotation about the $x$, $y$, or $z$ axis (commonly referred to as pitch, yaw, or roll).
For $\theta\in S^1$, we will refer to \emph{$\theta$ camera translation}, by which we mean translation of $\cos(\theta)$ units to the right, $\sin(\theta)$ units up, and no units inwards or outwards.
The most common camera translations are when $\theta=0$ or $\pi$, i.e.\ when the camera is translated to the left or right, for example if the camera is mounted on a horizontally moving car or held by a horizontally walking human~\cite{roth2007spatial}.

Adams and Carlsson~\cite{Range} show that high-contrast range patches are dense near the range patch primary circle $\{\cos(\alpha)e_1+\sin(\alpha)e_2\ |\ \alpha\in S^1\}$ (see Figure~\ref{fig:primBin}(a)).
Consider pairing the common horizontal ($\theta=0$ or $\pi$) camera translations with primary circle range patches.
Under camera translation in the $xy$ plane, the flow vector at a foreground pixel has the same direction but greater magnitude than at a background pixel.
After the mean-centering normalization in Step~4 of Section~\ref{sec:space}, a $\theta=0$ camera translation over the range patch $\cos(\alpha)e_1+\sin(\alpha)e_2$ produces the optical flow patch $\cos(\alpha)e_1^u+\sin(\alpha)e_2^u$, and a $\theta=\pi$ camera translation produces flow patch $-\cos(\alpha)e_1^u-\sin(\alpha)e_2^u$.
Hence when applied to all primary circle range patches, $\theta=0$ or $\pi$ translations produce the horizontal flow circle $\{\cos(\alpha)e_1^u+\sin(\alpha)e_2^u\ |\ \alpha\in S^1\}$ in Figure~\ref{fig:gk300c30}.

\subsection{Fiber bundles}\label{ssec:fiber}

Our torus model for the MPI-\emph{Sintel} optical flow dataset is closely related to the notion of a fiber bundle.
A fiber bundle is a tuple $(E,B,f,F)$, where $E$, $B$, and $F$ are topological spaces, and where $f\colon E\to B$ is a continuous map satisfying the so-called \textit{local triviality} condition.
Space $B$ is the \emph{base space}, $E$ is the \emph{total space}, and $F$ is the \emph{fiber}.
The local triviality condition states that given $b\in B$, there exists an open set $U\subseteq B$ containing $b$ and a homeomorphism $\varphi\colon f^{-1}(U)\to U\times F$ such that $\text{proj}_{U}\circ \varphi=f|_{f^{-1}(U)}$, where $\text{proj}_U$ denotes the projection onto the $U$--component.
In other words, we require $f^{-1}(U)$ to be homeomorphic to $U\times F$ in a consistent fashion.
Therefore, for any $p\in B$, we have $f^{-1}(\{p\})\cong F$.
Locally, the total space $E$ looks like $B\times F$, whereas globally, the different copies of the fiber $F$ may be ``twisted'' together to form $E$.

Both the cylinder and the M\"{o}bius band are fiber bundles with base space the circle $S^1$, and with fibers the unit interval $[0,1]$.
The cylinder is the product space $S^1\times [0,1]$, whereas in the M\"{o}bius band, the global structure encodes a ``half twist'' as one loops around the circle.
Locally, however, both spaces look the same, as each have the same fiber above each point of $S^1$.
The torus and the Klein bottle similarly are each fiber bundles over $S^1$, with fibers also $S^1$; indeed, they are the only two circle bundles over the circle.
Here the torus is the product space $S^1\times S^1$, whereas the Klein bottle is again ``twisted."

We use persistent homology in Section~\ref{ssec:tor} to justify a model for the MPI-\emph{Sintel} dataset that is naturally equipped with the structure of a fiber bundle over a circle, with each fiber being a circle.
A priori it is not clear whether this fiber bundle model should be the orientable torus or the nonorientable Klein bottle (which do occur in nature, as in the space of optical image patches~\cite{carlsson2008local}).
Via an orientation check, we will furthermore show that this optical flow fiber bundle model is a torus.

\subsection{A torus model for optical flow}\label{ssec:tor} 

We now describe the torus model for high-contrast patches of optical flow.
Define the map $f:S^1\times S^1 \mapsto \mathbb{R}^{18}$ via
\begin{equation*}
f(\alpha,\theta)=\cos(\theta)\Bigl(\cos(\alpha)e_1^u+\sin(\alpha)e_2^u\Bigr)+\sin(\theta)\Bigl(\cos(\alpha)e_1^v+\sin(\alpha)e_2^v\Bigr).
\end{equation*}
For $(\alpha,\theta)\in S^1\times S^1$, this means that $f(\alpha,\theta)$ is the optical flow patch produced from $\theta$ camera translation over the primary circle range patch $\cos(\alpha)e_1+\sin(\alpha)e_2$.

One obtains the horizontal flow circle by restricting to common camera motions $\theta\in\{0,\pi\}$, and by allowing the range patch parameter $\alpha\in S^1$ to be arbitrary.
When both parameters are allowed to vary over $S^1$, we hypothesize that a larger model for flow patches is produced.
Therefore, we ask: What is the image space $\im(f)$, i.e., what space do we get when both inputs $\alpha$ and $\theta$ are varied?

\begin{figure}[htp]
\centering
\includegraphics[scale=0.13]{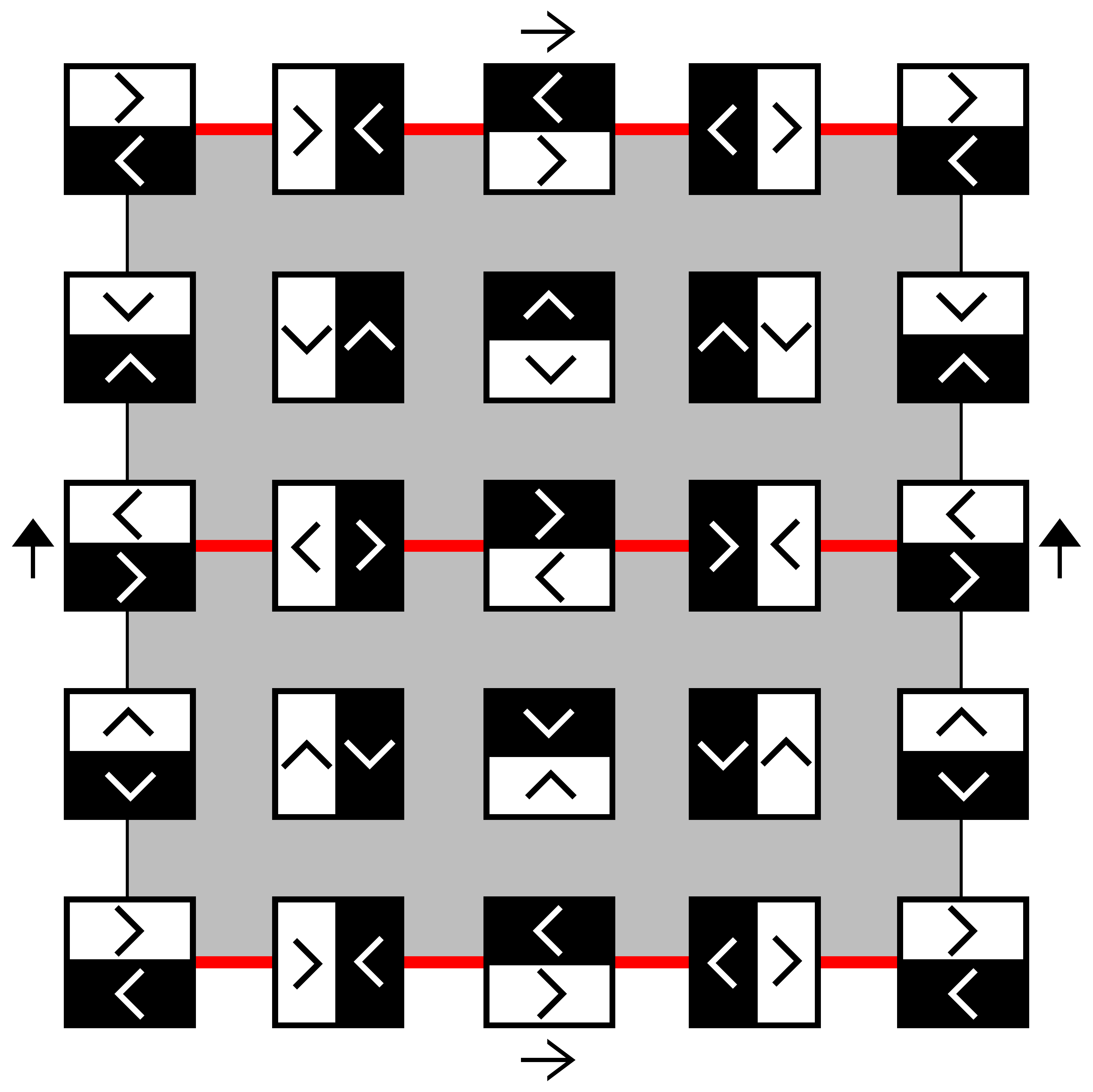}\\
\includegraphics[scale=0.13]{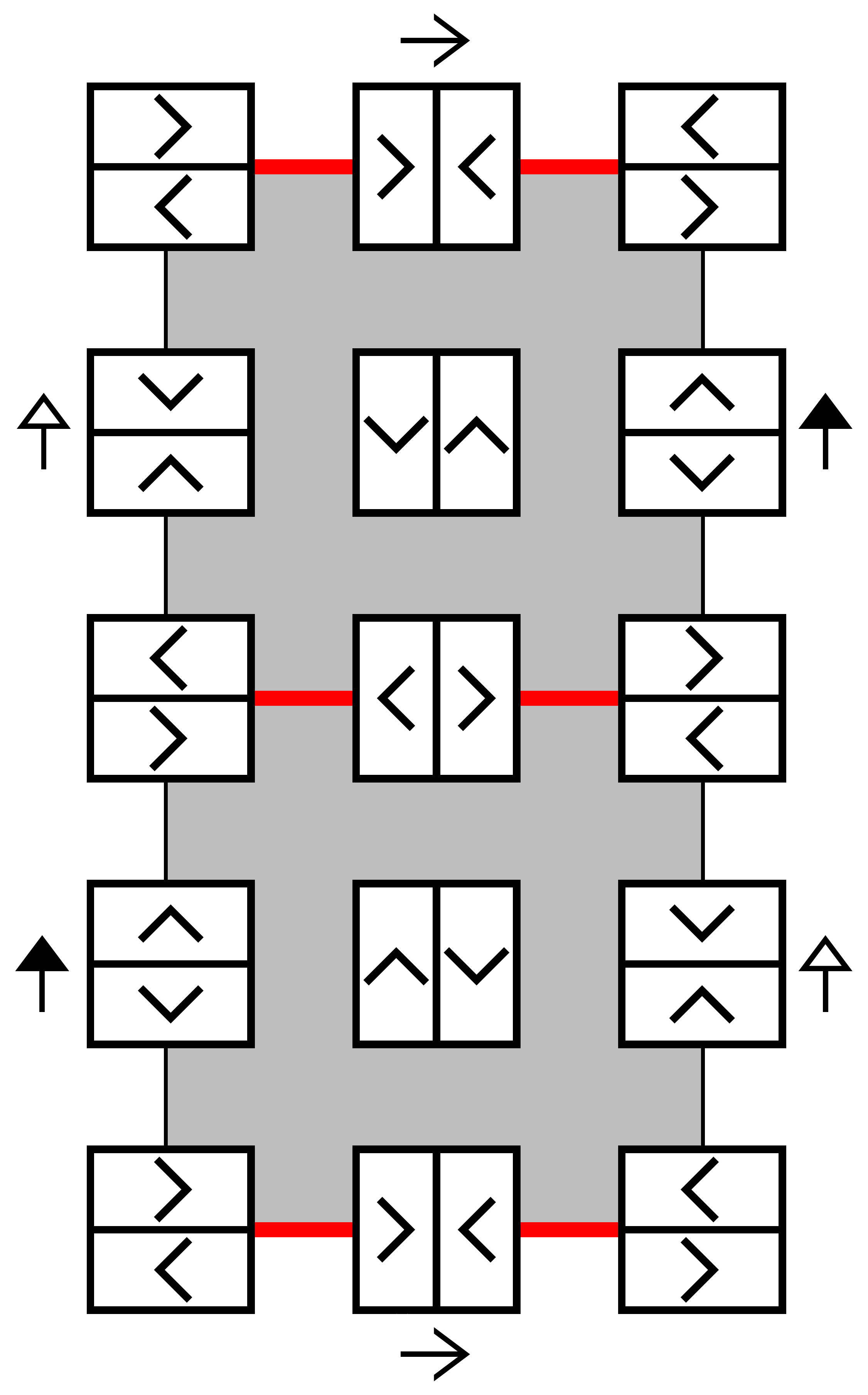}
\hspace{5mm}
\includegraphics[scale=0.13]{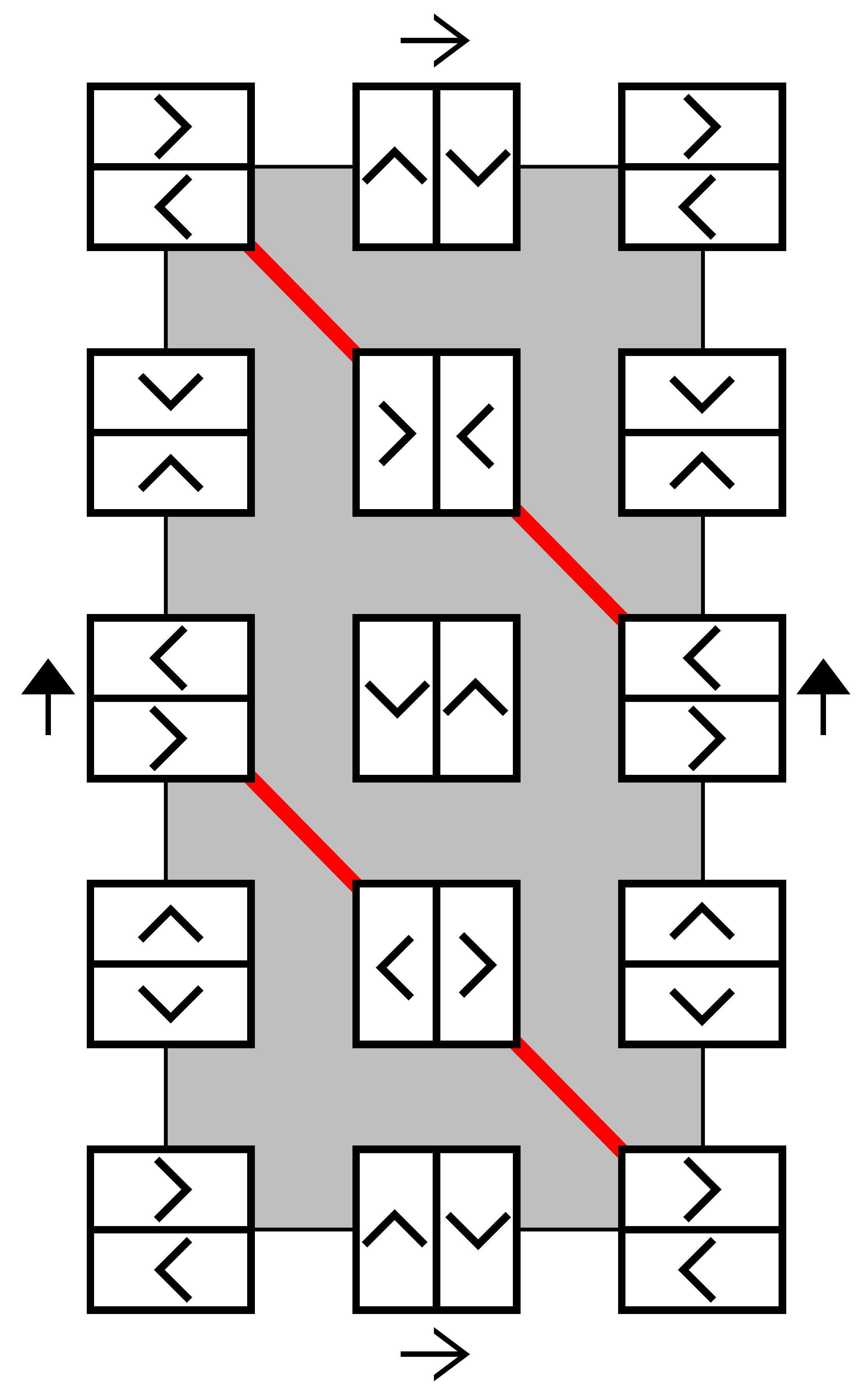}
\caption{(Top) The domain of $f$, namely $\{(\alpha,\theta)\in S^1\times S^1\}$.
(Bottom) The flow torus $\im(f)$.
The horizontal axis is the angle $\alpha$, and the vertical axis is the angle $\theta$ (respectively $\theta-\alpha$ on the right).
The horizontal flow circle is in red.}
\label{fig:tor}
\end{figure}

\begin{wrapfigure}{r}{0.3in}
\centering
\vspace{-\intextsep}
\includegraphics[width=0.3in]{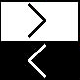}
\vspace{-\intextsep}
\end{wrapfigure}
Figure~\ref{fig:tor}~(top) shows the domain of $f$, namely $\{(\alpha,\theta)\in S^1\times S^1\}$.
This space, obtained by identifying the outside edges of the square as indicated by the arrows, is a torus.
In the insert to the right we show a sample patch on this torus.
The black and white rectangles are the foreground and background regions, respectively, of the underlying range patch.
The angle of the line separating the the foreground and background region is given by the parameter $\alpha$.
The direction $\theta$ of camera translation is given by the black arrow ($>$, $\vee$, $<$, or $\wedge$) in the white foreground rectangle.
The black and white arrows together show the induced optical flow vector field $f(\alpha,\theta)$.
In Figure~\ref{fig:tor}~(top), parameter $\alpha$ varies in the horizontal direction, and parameter $\theta$ varies in the vertical direction.

For two points $(\alpha,\theta),(\alpha',\theta')\in S^1\times S^1$, we have
\[f(\alpha,\theta)=f(\alpha',\theta') \Leftrightarrow(\alpha,\theta)=(\alpha',\theta') \mbox{ or } (\alpha,\theta)=(-\alpha',-\theta').\]
This means that under the map $f$, antipodal points in Figure~\ref{fig:tor}~(top) produce the same flow patch.
For instance, the horizontal flow circle appears twice in red (note the top and bottom edges of the square are identified).
It follows that the image space $\im(f)$ is homeomorphic to the quotient space $\{(\alpha,\theta)\in S^1\times S^1\} / \sim$, where $\sim$ denotes the identification $(\alpha,\theta)\sim(-\alpha,-\theta)$.
A torus with antipodal points identified remains a torus, and we refer to $\im(f)$ as the {\em optical flow torus}; see Figure~\ref{fig:tor}~(bottom).
The right and left edges of the bottom right image are identified by shifting one upwards by half its length (not by twisting) before gluing, which suggests a change of coordinates.
In Figure~\ref{fig:tor}~(bottom right) we plot the same flow torus, except that we replace the vertical parameter $\theta$ with $\theta-\alpha$.
The horizontal flow circle in red now wraps once around one circular direction, and twice around the other direction.

\begin{figure}[htb]
\centering
\includegraphics[scale=0.12]{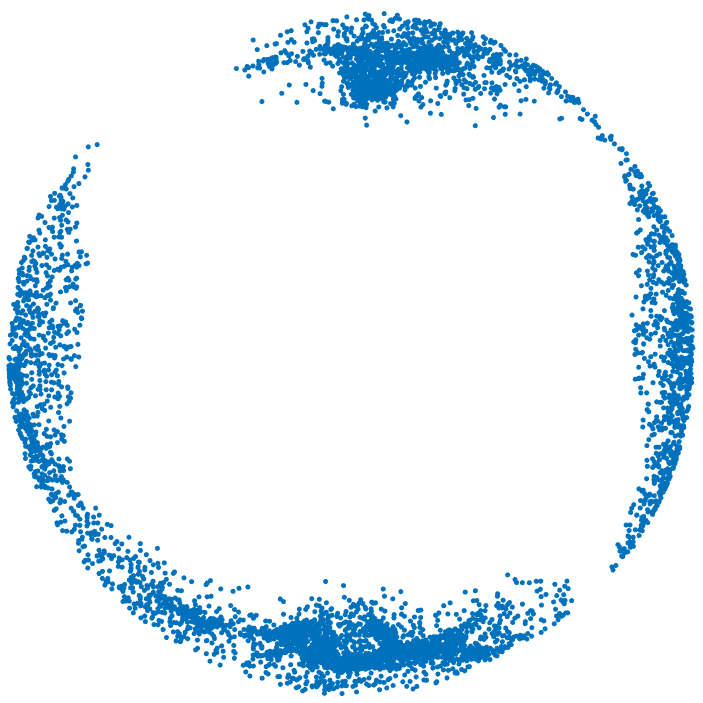}
\hspace{5mm}
\includegraphics[scale=0.12]{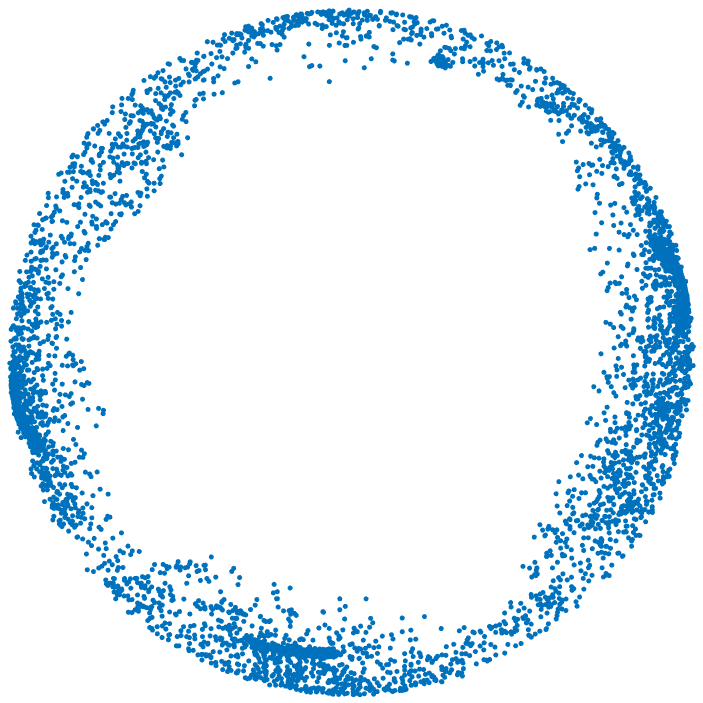}
\hspace{5mm}
\includegraphics[scale=0.12]{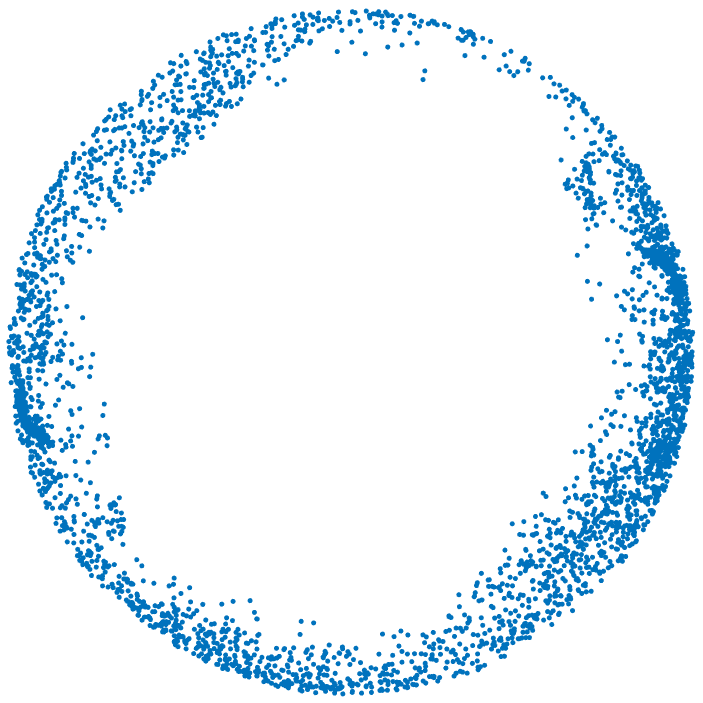}\\
\vspace{5mm}
\includegraphics[scale=0.12]{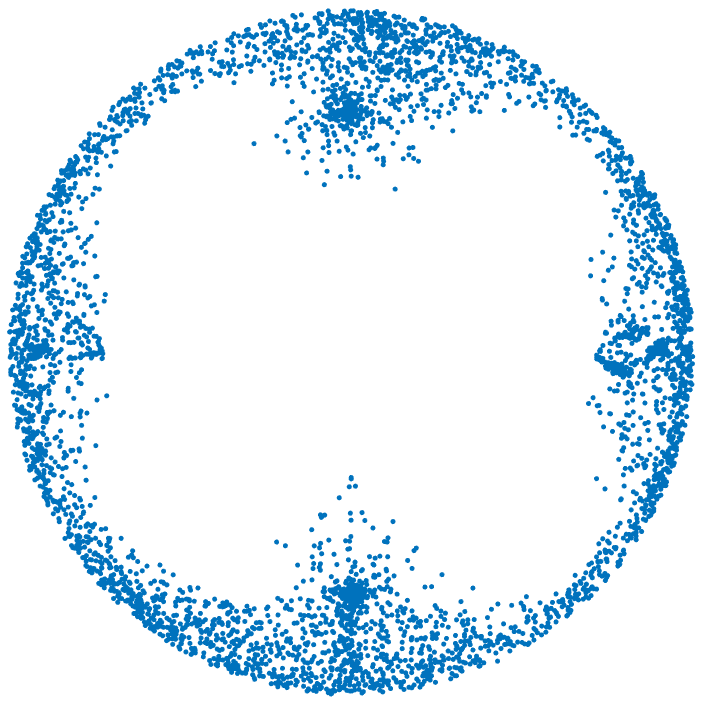}
\hspace{5mm}
\includegraphics[scale=0.12]{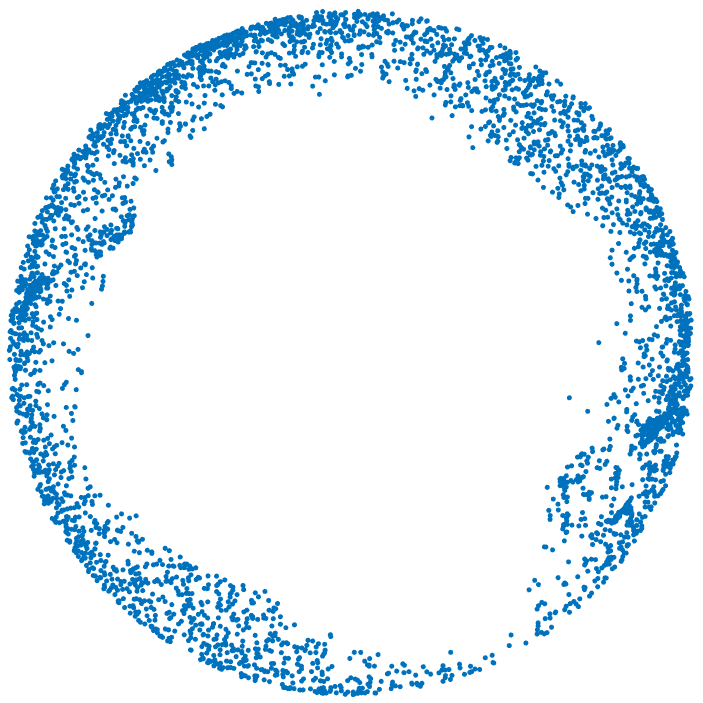}
\hspace{5mm}
\includegraphics[scale=0.12]{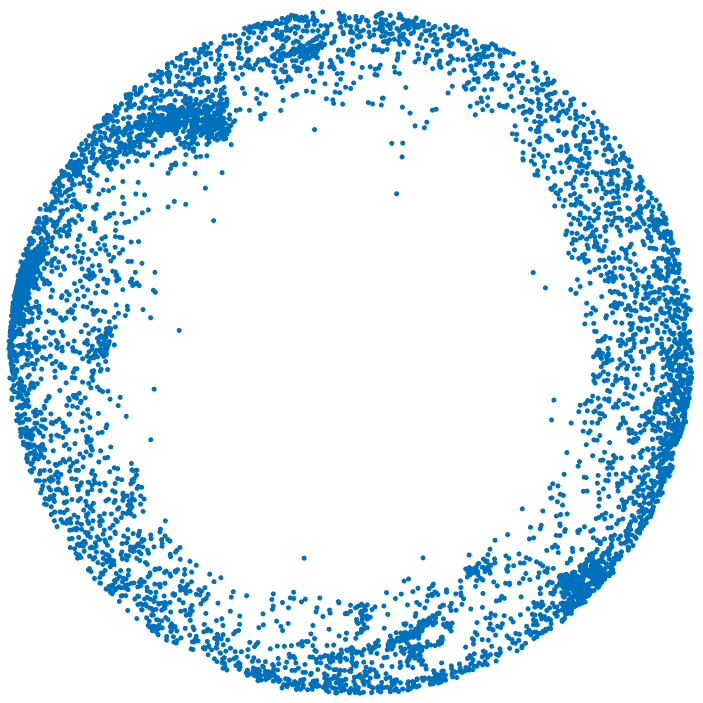}
\caption{The set of patches in $X_\theta(300,50)$, projected to the plane spanned by the basis vectors $\cos(\theta)e_1^u+\sin(\theta)e_1^v$ (horizontal axis) and $\cos(\theta)e_2^u+\sin(\theta)e_2^v$ (vertical axis).
From left to right, we show $\theta=0,\frac{\pi}{6},\frac{2\pi}{6},\frac{3\pi}{6},\frac{4\pi}{6},\frac{5\pi}{6}$.
The projected circles together group together to form a torus.
}
\label{fig:torus_projections}
\end{figure}

\begin{figure}[htb]
\centering
\includegraphics[height=1.6in]{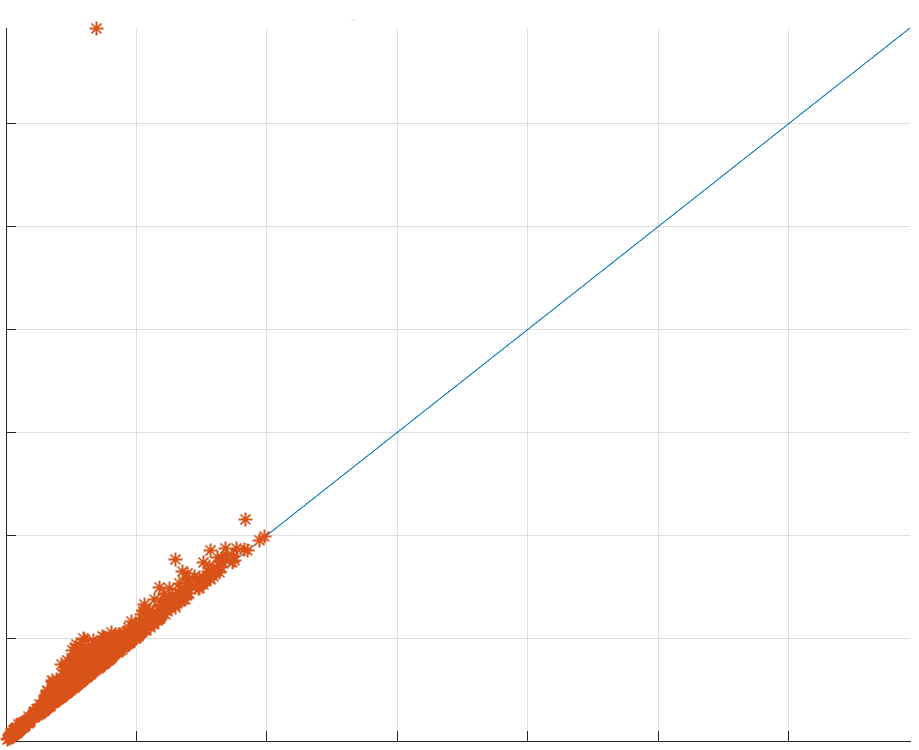}
\hspace{10mm}
\includegraphics[height=1.6in]{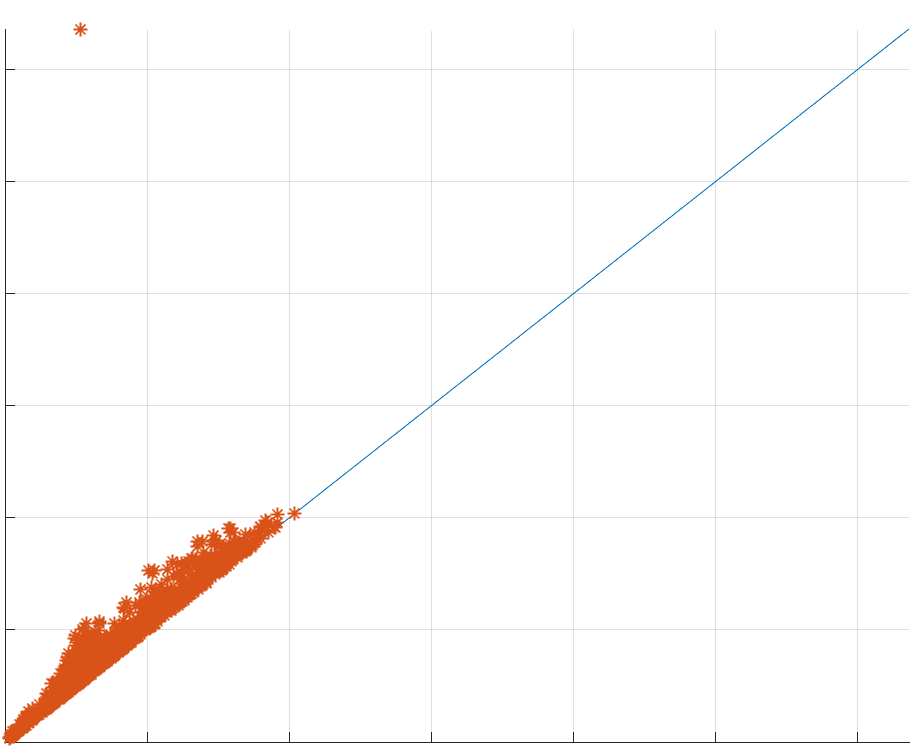}
\caption{The 1-dimensional persistent homology of Vietoris--Rips complexes of $X_\theta(300,30)$, computed in Ripser~\cite{bauer2017ripser}, illustrate that these data sets are well-modeled by circles (one significant 1-dimensional feature in the top left of each plot).
Persistence diagrams contain the same content as persistence intervals, just in a different format: each point is a topological feature with birth scale and death scale given by its $(x,y)$ coordinates.
We plot two sample angles: $\theta=\frac{3\pi}{12}$ (left) and $\theta=\frac{7\pi}{12}$ (right).}
\label{fig:torus_fiber_persistence}
\end{figure}

We provide experimental evidence that $\im(f)$, the optical flow torus, is a good model for high-contrast optical flow.
Figures~\ref{fig:torus_projections} and~\ref{fig:torus_fiber_persistence} show that for any angle $\theta$, the patches $X_\theta(300,30)$ (with predominant flow in direction $\theta$) form a circle.
These circles group together to form a torus, which furthermore has a natural fiber bundle structure.
Indeed, the map from the torus to the predominant angle $\theta$ of each patch is a fiber bundle, whose total space is a torus, whose base space is the circle of all possible predominant angles $\theta$, and whose fibers are circles (arising from the range image primary circle in Figure~\ref{fig:primBin}).

\begin{figure}[htb]
\centering
\includegraphics[width=3.5in]{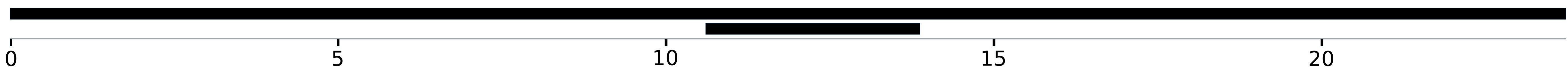}
\caption{A 1-dimensional zigzag persistence computation, showing that the circles in Figure~\ref{fig:torus_projections} glue together with the structure of a fiber bundle.
The 24 horizontal steps correspond to the 12 spaces $X_\theta(300,50)$ and the 12 intermediate spaces which are unions thereof.}
\label{fig:zigzag}
\end{figure}

To confirm that the circular fibers glue together to form a fiber bundle, we do a zigzag persistence computation.
\begin{center}
\begin{tikzpicture}
\node at (-0.5, 0.45) (a) {$X_0(300,50)$};
\node at (1, -0.45) (b) {$X_0(300,50) \cup X_\frac{\pi}{12}(300,50)$};
\node at (2.5, 0.45) (c) {$X_\frac{\pi}{12}(300,50)$};
\node at (4, -0.45) (d) {$\cdots$};
\node at (5.5, 0.45) (e){$X_\frac{11\pi}{12}(300,50)$};
\draw [right hook ->] (a) -- (b);
\draw [left hook ->] (c) -- (b);
\draw [right hook ->] (c) -- (d);
\draw [left hook ->] (e) -- (d);
\end{tikzpicture}
\end{center}
Figure~\ref{fig:zigzag} shows the one-dimensional zigzag persistence of Vietoris--Rips complexes built on top of the following zigzag diagram, confirming that the circles piece together compatibly into a fiber bundle structure.

In more detail, for twelve different angle bins $\theta\in\{0,\frac{\pi}{12},\ldots\frac{11\pi}{12}\}$ we construct the dense core subsets $X_{\theta}(300,50)$.
For computational feasibility, we then downsample via sequential maxmin sampling~\cite{de2004topological} to reduce each set $X_{\theta}(300,50)$ to a subset of $50$ data points; Ripser computations show the persistent homology is robust with regard to this downsampling.
After building a zigzag filtration as described above, we use Dionysus~\cite{dionysus} to compute the zigzag homology barcodes in Figure~\ref{fig:zigzag}.
The single long interval confirms that the circles indeed piece together compatibly.

\begin{figure}[htb]
\centering
\includegraphics[width=3.8in]{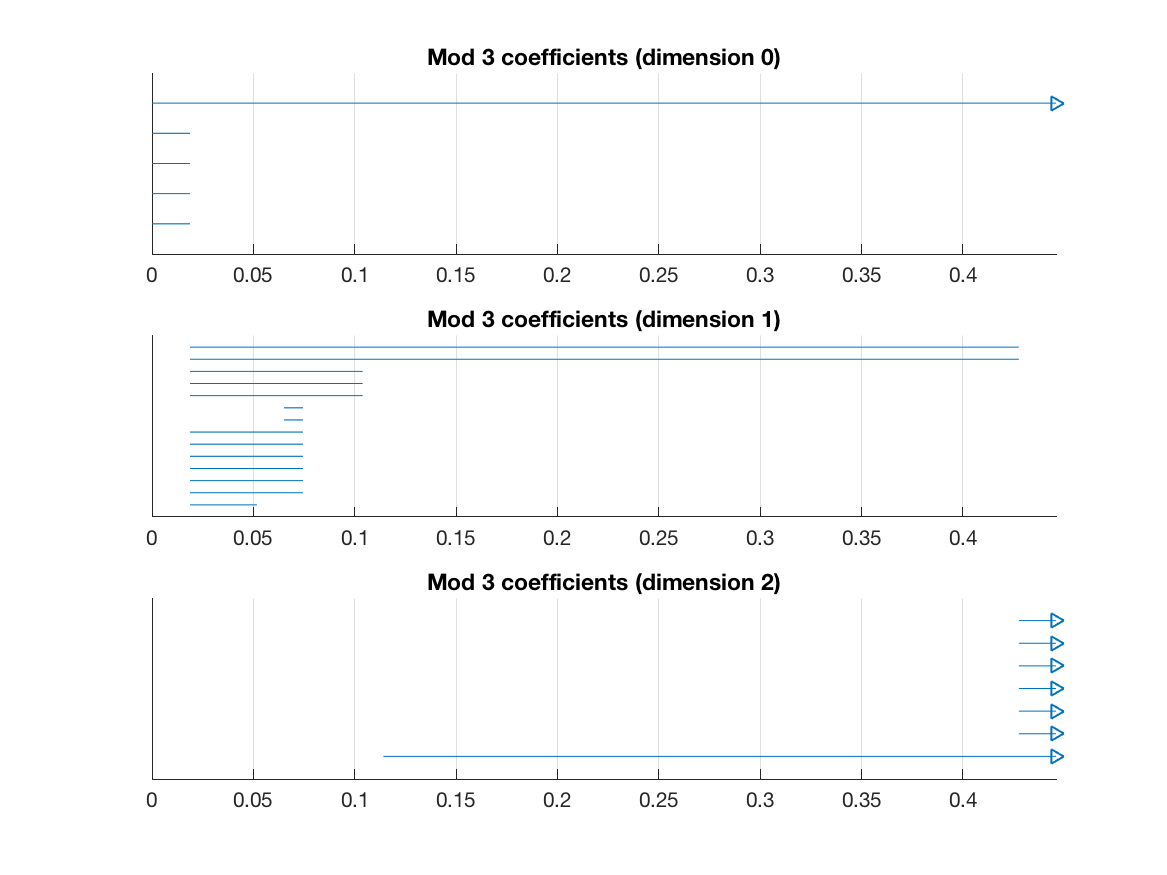}
\caption{With $\mathbb{Z}/3\mathbb{Z}$ coefficients, the Betti number signature $\beta_0=1$, $\beta_1=2$, $\beta_2=1$ identifies the model as a torus instead of a Klein bottle.}
\label{fig:orientation}
\end{figure}

We have experimentally verified a fiber bundle model with base space a circle and with fiber a circle.
As the only circle bundles over the circle are the torus and the Klein bottle, this rules out many possible shapes for our model --- our model can no longer be (for example) a sphere, a double torus, a triple torus, a projective plane, etc.
It remains to identify this fiber bundle model as either the torus or the Klein bottle.
One test would be to check if the orientation on a generator for the 1-dimensional homology of $X_0(300,50)$ is preserved after looping once around the circle; generator orientation would be preserved for a torus but not the Klein bottle.
Another way to verify that this fiber bundle is a torus instead of a Klein bottle could be to use persistence for circle-valued maps~\cite{burghelea2013topological}, on the map from the total space to the circle that encodes the predominant angle $\theta$ of each flow patch.
We instead take a computational approach that does not require tracking generators.
Indeed, we sample a collection of patches near the idealized optical flow torus, and compute the persistent homology of a witness complex construction both with $\mathbb{Z}/2\mathbb{Z}$ and $\mathbb{Z}/3\mathbb{Z}$ coefficients (Figure~\ref{fig:orientation}).
We obtain the Betti signature $\beta_0=1$, $\beta_1=2$, $\beta_2=1$ with both choices of coefficients, confirming that we indeed have a torus.
Indeed, a Klein bottle with $\mathbb{Z}/3\mathbb{Z}$ coefficients would instead lose one long bar in both homological dimensions one and two, giving $\beta_0=1$, $\beta_1=1$, $\beta_2=0$ with $\mathbb{Z}/3\mathbb{Z}$ coefficients.

The 2-dimensional flow torus model does not model all common optical flow patches; for example it omits patches arising from zooming in, zooming out, or roll camera motions.

\section{Conclusions}\label{sec:con}
We explore the nonlinear statistics of high-contrast $3\times 3$ optical flow patches from the computer-generated video short \emph{Sintel} using topological machinery, primarily persistent homology and zigzag persistence.
Since no instrument can measure ground-truth optical flow, an understanding of the nonlinear statistics of flow is needed in order to serve as a prior for optical flow estimation algorithms.
With a global estimate of density, the densest patches lie near a circle, the horizontal flow circle.
Upon selecting the optical flow patches whose predominant direction of flow lies in a small bin of angle values, we find that the patches in each such bin are well-modeled by a circle.
By combining these bins together we obtain a torus model for optical flow, which furthermore is naturally equipped with the structure of a fiber bundle, over a circular base space of common range image patches.

\section{Acknowledgements}
We thank Gunnar Carlsson, Bradley Nelson, Jose Perea, and Guillermo Sapiro for helpful conversations.

\bibliographystyle{plain}
\bibliography{flow}

\end{document}